\documentclass[11pt]{article}

\usepackage[preprint]{acl}

\usepackage{times}
\usepackage{latexsym}
\usepackage[table]{xcolor}
\usepackage{tabularx}  

\usepackage[T1]{fontenc}

\usepackage[utf8]{inputenc}
\usepackage{subfigure}
\usepackage{enumitem}
\usepackage{stfloats}
\usepackage{mdframed}
\usepackage{listings}
\usepackage{booktabs} 
\usepackage{hyperref}
\usepackage{amsmath}
\usepackage{amssymb}
\usepackage{mathtools}
\usepackage{amsthm}
\usepackage{multirow}
\usepackage{array}
\usepackage{cuted}
\usepackage{capt-of}
\usepackage{multicol}
\usepackage{graphicx} 

\usepackage[ruled]{algorithm2e}
\usepackage{silence}
\usepackage{float}   

\usepackage{microtype}
\lstdefinestyle{inlinecode}{
  basicstyle=\ttfamily\small,
  breaklines=true,
  breakatwhitespace=true,
  columns=fullflexible,
  keepspaces=true,
  frame=none,
  aboveskip=0pt,
  belowskip=0pt,
}
 
\mdfdefinestyle{subframe}{
  linewidth=0.5pt,
  linecolor=black!30,
  backgroundcolor=black!3,
  innerleftmargin=8pt,
  innerrightmargin=8pt,
  innertopmargin=6pt,
  innerbottommargin=6pt,
  skipabove=4pt,
  skipbelow=4pt,
  roundcorner=3pt,
}
 
\mdfdefinestyle{outerframe}{
  linewidth=0.8pt,
  linecolor=black!50,
  backgroundcolor=white,
  innerleftmargin=10pt,
  innerrightmargin=10pt,
  innertopmargin=8pt,
  innerbottommargin=8pt,
  skipabove=0pt,
  skipbelow=0pt,
}

\usepackage{colortbl}
\usepackage{pifont}
\usepackage{makecell}

\definecolor{lightblue}{RGB}{235,243,255}
\definecolor{headerblue}{RGB}{220,232,250}
\definecolor{midgray}{RGB}{120,120,120}

\newcommand{\best}[1]{\textbf{#1}}
\newcommand{\second}[1]{\underline{#1}}
\newcommand{\na}{\textcolor{midgray}{--}}

\definecolor{sevenclr}{RGB}{0,100,210}

\newcommand{\sd}[1]{$_{\pm\text{#1}}$}

\usepackage{inconsolata}

\title{Clearing the FoG: Training KG Agents \\ to Retrieve the Evidence Behind Their Answers}

\author{
  \textbf{Zongyuan Sun\textsuperscript{1,*}},
  \textbf{Shuai Zhao\textsuperscript{1,$\dagger$}},
  \textbf{Ke Shu\textsuperscript{2,*}},
  \textbf{Yingjun Du\textsuperscript{3}},
  \textbf{Ziyue Zhu\textsuperscript{1}},
  \textbf{Haoran Luo\textsuperscript{4,$\dagger$}}
\\
  \textsuperscript{1}Beijing University of Posts and Telecommunications, Beijing, China \\
  \textsuperscript{2}University of Helsinki, Helsinki, Finland \quad
  \textsuperscript{3}University of Amsterdam, Amsterdam, Netherlands \\
  \textsuperscript{4}Nanyang Technological University, Singapore
\\
  \small{\textsuperscript{*}Equal contribution.\quad \textsuperscript{$\dagger$}Corresponding authors.}
}

\begin{document}
\maketitle

\begin{abstract}
Large language models (LLMs) can answer many factual questions correctly, but a correct answer is not necessarily a supported one. For an LLM agent, a knowledge graph (KG) plays two roles: as a source of answers, it supplies facts that the model lacks; as a source of evidence, it supplies the supporting triples that make an answer traceable and auditable. However, existing KG agents treat the KG only as a source of answers: their training data carries answers but not evidence, and their training rewards check the answer but not the retrieved evidence, so agents often answer correctly without retrieving the supporting triples. We propose \textbf{FoG}, a framework that trains KG agents to retrieve the evidence behind their answers. FoG consists of evidence-grounded synthesis, a KG interaction agent, and an evidence-coverage process reward (ECPR): the synthesis recursively hides explicit entities behind verifiable sub-questions, so that every question carries its supporting triples by construction; the agent records the triples returned by the KG; and ECPR credits each supporting triple to the turn that first retrieves it. Experiments on three KGQA benchmarks demonstrate that FoG enables 3B--4B open-source agents to outperform substantially larger baselines in answer accuracy and evidence coverage, with zero-shot cross-KG transfer.
\end{abstract}

\section{Introduction}

\begin{figure}[t]
    \centering
    \includegraphics[width=1\linewidth]{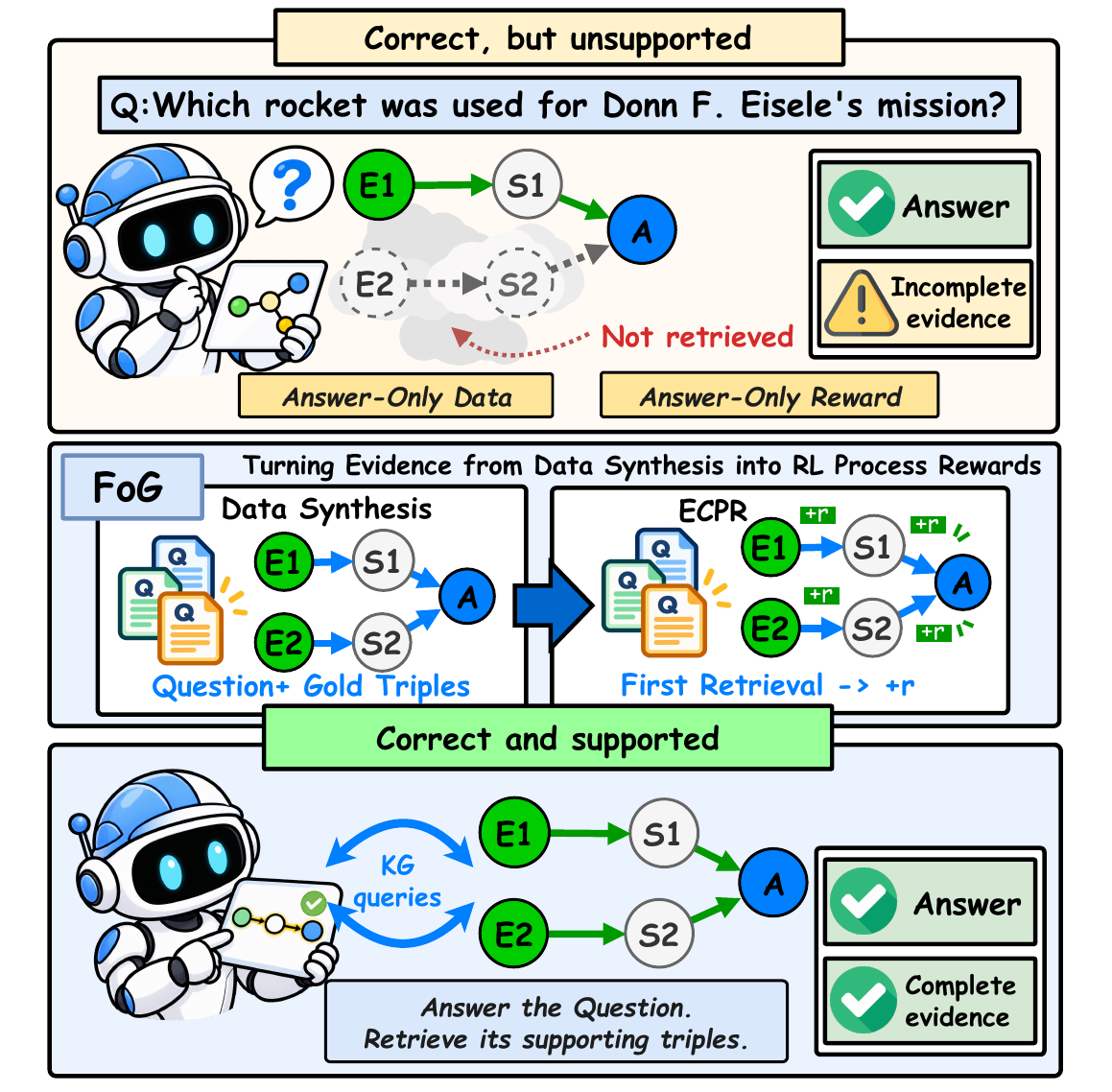}
    \caption{\textbf{Motivation of FoG.}
    Answer-only training yields correct but unsupported answers, whereas FoG embeds supporting triples at synthesis and rewards their retrieval during RL training.}
    \label{fig:intro}
    \vspace{-5mm}
\end{figure}

Large language models (LLMs)~\citep{kimi,GLM,qwen,llama,gpt-4} increasingly answer factual questions from parametric knowledge, but remain difficult to trust on questions that require \emph{complex factual reasoning}~\cite{mirzadeh2025gsm,hu2024large,jullien2025knowledge,petroni2021kilt}. A correct answer is not necessarily a supported one: the model may answer from memory or surface cues while exposing little about \emph{why} the answer holds~\citep{hallucination_1,hallucination_2,nguyen2024direct}, which is problematic in high-stakes settings such as healthcare and law~\cite{cabello2024meg}, where users need evidence that they can check~\cite{sui2025fidelis,agrawal2024mindful}. Current LLM answers thus lie in a \emph{fog}: the answer is visible, yet the evidence behind it remains hidden.

\begin{figure*}[t]
    \centering
    \includegraphics[width=1\textwidth]{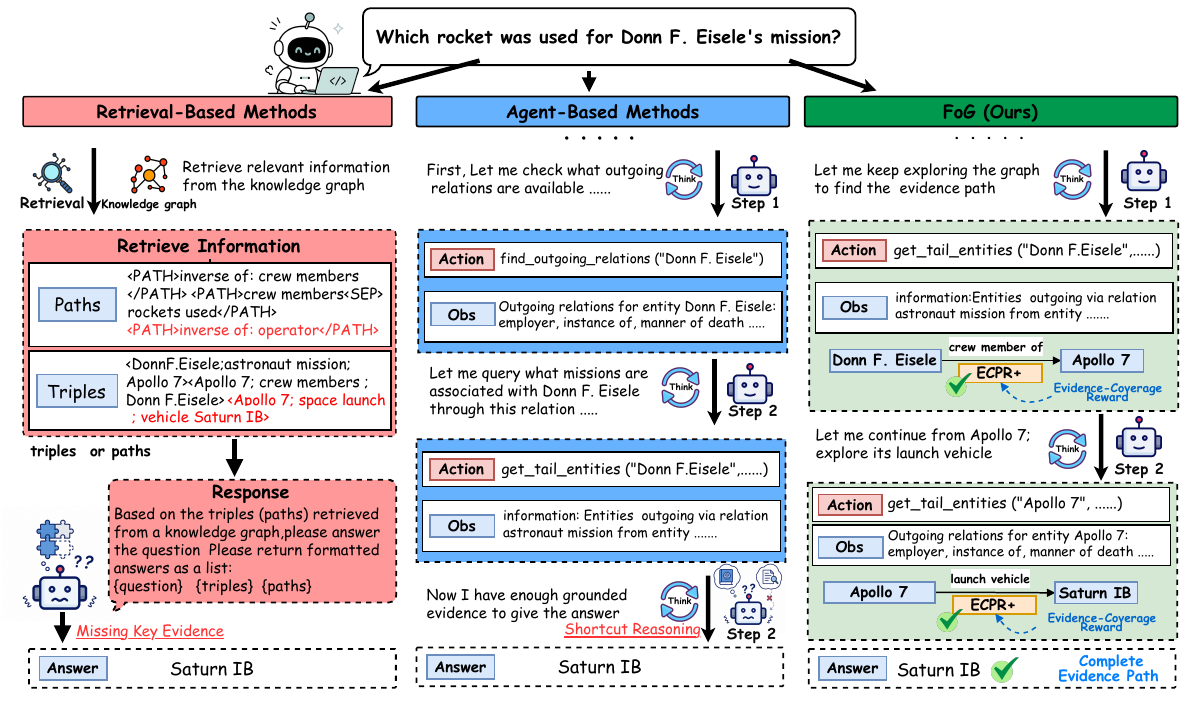}
    \caption{\textbf{Comparison of KG-augmented reasoning paradigms.}
    All three methods return the correct answer, but retrieval-based methods miss key evidence and agent-based methods answer after the first hop, whereas FoG retrieves every supporting triple of the answer.}
    \label{fig:related_work}
    \vspace{-3mm}
\end{figure*}

Knowledge graphs (KGs)~\cite{roadmap} offer a natural way to clear this fog, since they store facts as verifiable triples. Existing KG-augmented methods follow two paradigms (Figure~\ref{fig:related_work}). \emph{Retrieval-based} methods~\citep{RoG,SubGraphRAG,GCR} retrieve triples, paths, or subgraphs and answer from them. \emph{Agent-based} methods~\citep{ToG,PoG,KG-R1,kg-agent} interact with the KG through a multi-turn ``think--act--observe'' process, and recent work trains them with reinforcement learning (RL)~\citep{KG-R1,KnowCoder-A1,KBQA-R1,SoG-R1}.

However, existing KG agents treat the KG only as a \textbf{source of answers} and overlook its role as a \textbf{source of evidence}. An LLM with no KG access can already answer many benchmark questions correctly, and even an RL-trained agent can answer correctly while retrieving only part of the supporting triples. We identify two causes. First, \emph{training data carries answers but not evidence}: most KGQA datasets annotate answers and at most logical forms~\cite{cwq,webqsp,grailqa}, and recovering the supporting triples afterwards requires enumerating and verifying paths for every question. Second, \emph{training rewards check the answer but not the evidence}: most rewards score only the final answer or whether the KG returns it~\cite{KG-R1,Plan-Then-retrieve,GraphWalker,KnowCoder-A1,KBQA-R1}, so a trajectory that reaches the answer through part of the evidence earns as much as one that retrieves all of it. The process rewards that go further depend on an LLM judge or on gold paths recovered for every question~\cite{KG-Hopper,EoG}, both of which are costly. Both causes trace back to one bottleneck: \emph{evidence is expensive to recover once a question has been written}.

To remove this bottleneck, we propose \textbf{FoG} (Figure~\ref{fig:intro}), a framework that trains KG agents to retrieve the evidence behind their answers. Our key observation is that evidence is hard to recover after a question is written but can be embedded when the question is constructed and then rewarded during training. First, \textbf{evidence-grounded synthesis} recursively hides the explicit entities of a seed question behind verifiable sub-questions, so every question carries its supporting triples. Second, a \textbf{KG interaction agent} answers through multi-turn KG queries and records every triple that the KG returns. Lastly, during RL training, an \textbf{evidence-coverage process reward} (ECPR) gives a dense reward to each turn that first retrieves one of these supporting triples. Our contributions are as follows:
\begin{itemize}[leftmargin=*,itemsep=0pt,topsep=2pt]
    \item We distinguish two roles of the KG and show that existing training data and rewards serve it only as a source of answers, not of evidence.
    \item We propose FoG, which embeds supporting triples into synthetic questions and rewards their retrieval through ECPR, so the dense reward comes from data synthesis with no LLM judge or extra annotation.
    \item On three benchmarks, 3B--4B FoG agents outperform larger baselines in accuracy and evidence coverage, and transfer zero-shot across KGs.
\end{itemize}

\section{Related Work}

\paragraph{Knowledge graph augmented reasoning.}
Unlike prompting strategies~\citep{CoT,Self-Consistency,ToT,GoT}, KG-augmented methods ground reasoning in verifiable triples and follow two paradigms. Retrieval-based methods~\citep{RoG,SubGraphRAG,GCR,GNN-RAG,G-Retrieval,PathMind} condition the LLM on triples, paths, or subgraphs retrieved in a single step. Agent-based methods~\citep{ToG,PoG,KG-R1,kg-agent} explore the KG iteratively: ToG and PoG prompt a frozen LLM, whereas recent work trains the agent with RL~\citep{KG-R1,Plan-Then-retrieve,KnowCoder-A1,KBQA-R1,KG-Reasoner,SoG-R1}. However, these agents use the KG only as a source of answers and can answer correctly without retrieving the supporting triples. In this work, we train agents to use it as a source of evidence.

\paragraph{Synthetic data for KGQA.}
Synthetic data frees KGQA from human annotation, since questions can be generated from the KG and verified by execution. It provides questions for agent training~\citep{GraphWalker} and for benchmarking and retriever training~\citep{KGQAGen,GTSQA}. However, these pipelines treat it as extra examples to train or evaluate on. In this work, we treat it as a source of supervision: every synthesized question carries the evidence that defines its reward.

\paragraph{Rewards for KG agents.}
RL lets a KG agent learn its retrieval policy from interaction with the KG. Most work trains from outcome rewards on the final answer with rule-based terms for valid interaction~\citep{KG-R1,Plan-Then-retrieve,KnowCoder-A1,KBQA-R1,KG-Reasoner,SoG-R1,Peak-Then-Collapse}. Because outcome rewards are sparse, later work adds dense process rewards, obtained from an LLM judge~\citep{KG-Hopper} or from gold paths and logical forms~\citep{EoG,RoE,GAPD}. However, both sources are costly: a judge must score every rollout, and gold references must be annotated or recovered for every question. In this work, the process reward comes from data synthesis: every synthesized question already carries its supporting triples, which provide a dense reward at no extra cost.

\section{Method}
\begin{figure*}[t]
    \centering
    \includegraphics[width=1.0\textwidth]{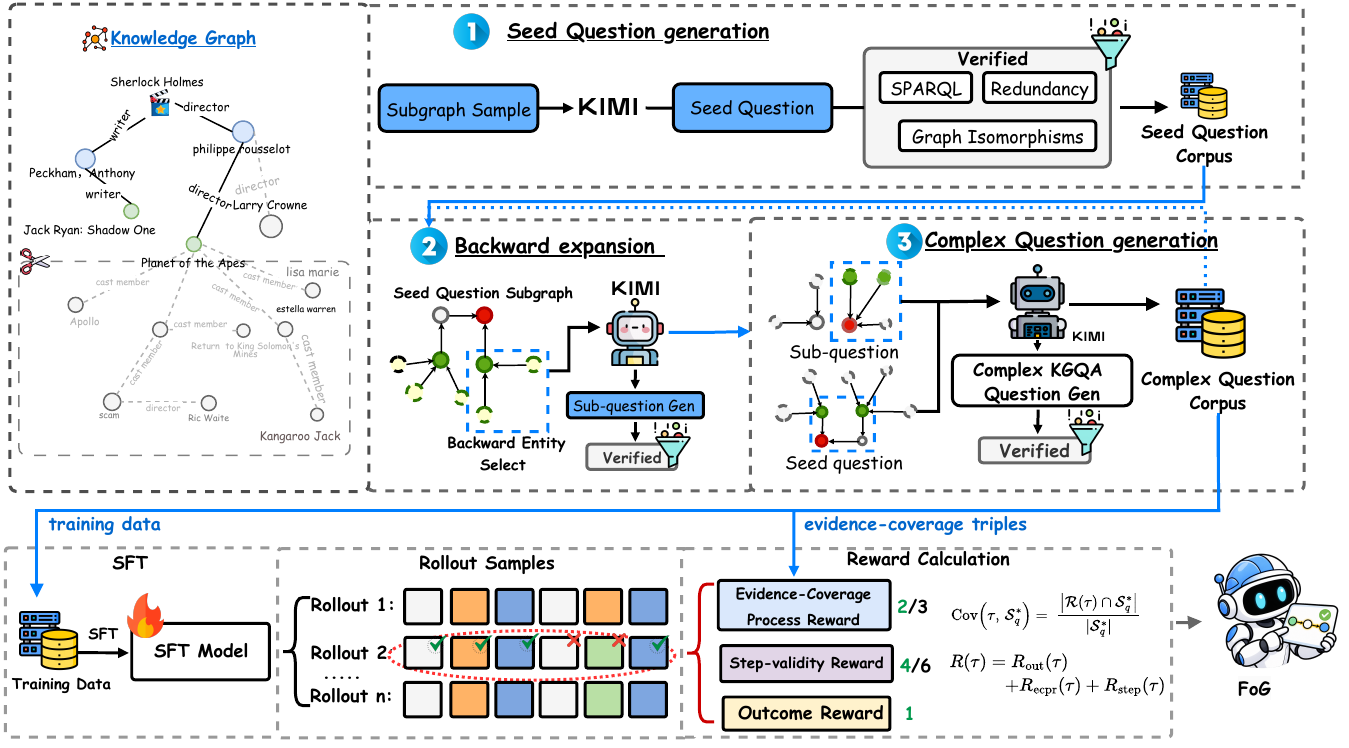}
    \caption{\textbf{The FoG pipeline.}
    Evidence-grounded synthesis produces complex questions with gold supporting triples, the KG interaction agent records the triples that it retrieves, and ECPR rewards the retrieval of the gold supporting triples.}
    \label{fig:methods}
    \vspace{-3mm}
\end{figure*}
In this section, we introduce FoG, as shown in Figure~\ref{fig:methods}, including evidence-grounded synthesis, a KG interaction agent, and ECPR.

\noindent\textbf{Definition 1 (KG question answering).} A knowledge graph is $\mathcal{G}=(\mathcal{E},\mathcal{R},\mathcal{T})$ with entities $\mathcal{E}$, relations $\mathcal{R}$, and triples $\mathcal{T}\subseteq \mathcal{E}\times\mathcal{R}\times\mathcal{E}$. A question $q$ has a logical form $\ell_q$ and a gold answer $y_q^{*}\in\operatorname{SPARQL}(\ell_q,\mathcal{T})$, where $\operatorname{SPARQL}(\ell,\mathcal{S})$ is the result of executing $\ell$ over a triple set $\mathcal{S}$. An agent answers $q$ through a trajectory $\tau = \bigl((s_1,a_1), \ldots, (s_T,a_T)\bigr)$ of states and actions and outputs a prediction $\hat{y}$, where $\mathcal{T}_t\subseteq\mathcal{T}$ are the triples returned at step $t$, $\mathcal{T}(\tau)=\bigcup_{t}\mathcal{T}_t$ all triples retrieved along $\tau$, and $\mathcal{E}(\tau)$ all returned entities.

\noindent\textbf{Definition 2 (Supported answer).} A triple set $\mathcal{S}\subseteq\mathcal{T}$ \emph{supports} the answer of $q$ if $y_q^{*}\in\operatorname{SPARQL}(\ell_q,\mathcal{S})$. The answer of $\tau$ is \emph{correct} if $\hat{y}$ matches $y_q^{*}$, \emph{reached} if $y_q^{*}\in\mathcal{E}(\tau)$, and \emph{supported} if $\mathcal{T}(\tau)$ supports it. Unlike the first two levels, which use the KG as a source of answers, a supported answer uses it as a source of evidence. In FoG, each training question carries \emph{gold supporting triples} $\mathcal{S}^{*}_{q}$ that support its answer.

\subsection{Evidence-Grounded Synthesis}
\label{sec:synthesis}

To obtain complex questions that carry their supporting evidence, we propose evidence-grounded synthesis that includes seed question construction, backward expansion, and quality control. All roles use the teacher Kimi-K2~\citep{kimi}.

\paragraph{Seed question construction.}
We sample a size-bounded subgraph $\mathcal{G}_{\text{sub}}$ from Wikidata around a seed entity $v_{\text{seed}}$ of moderate degree. The teacher generates a 1-hop question whose answer is $v_{\text{seed}}$:
\begin{equation}
\label{eq:seed}
    \bigl(Q^{(1)}, \ell^{(1)}, \mathcal{S}^{(1)}\bigr)
    \sim \pi_{\text{gen}}\bigl(v_{\text{seed}}, \mathcal{G}_{\text{sub}}\bigr),
\end{equation}
where $\pi_{\text{gen}}$ is the teacher model, $Q^{(1)}$ the seed question, $\ell^{(1)}$ its logical form, and $\mathcal{S}^{(1)}$ its supporting triples. The seed is retained only if every triple in $\mathcal{S}^{(1)}$ exists in $\mathcal{G}_{\text{sub}}$ and $\ell^{(1)}$ executes over the global KG with $v_{\text{seed}}$ in its result set.

\paragraph{Backward expansion.}
Given a verified question $Q^{(n)}$ with logical form $\ell^{(n)}$, answer $y=v_{\text{seed}}$, and supporting triples $\mathcal{S}^{(n)}$, we replace an explicit non-answer entity $u$ in $Q^{(n)}$ with a sub-question $\tilde{Q}$ whose answer is exactly $u$:
\begin{equation}
\begin{aligned}
    &Q^{(n+1)} = Q^{(n)}[u \rightsquigarrow \tilde{Q}], \\
    &\mathcal{S}^{(n+1)} = \mathcal{S}^{(n)} \cup \tilde{\mathcal{S}}, \\
    &y \in \operatorname{SPARQL}\bigl(\ell^{(n+1)}, \mathcal{S}^{(n+1)}\bigr),
\end{aligned}
\end{equation}
where $\tilde{\mathcal{S}}$ are the supporting triples of $\tilde{Q}$, drawn from the incoming neighborhood of $u$, and $\ell^{(n+1)}$ is the expanded logical form. Each expansion turns an explicit entity into a hidden intermediate variable, and the accumulated triples still support the answer. Thus $N\in\{0,1,2\}$ expansions yield 1- to 3-hop questions, where $N=0$ keeps the seed. Expansion uses a ReAct agent~\citep{react} that retrieves the neighborhood of $u$, selects triples, writes $\tilde{Q}$, and verifies it; on failure, it reselects triples or the entity.

\noindent\textbf{Proposition 1.} \textit{Backward expansion preserves the answer and yields triples that support it.}\vspace{-2mm}
\begin{proof}\renewcommand{\qedsymbol}{}
We provide experimental results in Section~\ref{sec:rq3} and proofs in Appendix~\ref{app:proof1}.
\end{proof}

\paragraph{Quality control.}
Each synthesized question must further pass two symbolic checks, which keep the answer specific and every topic entity necessary.
\textbf{(a) Executability.} Let $\mathcal{Y}^{(n)} = \operatorname{SPARQL}(\ell^{(n)}, \mathcal{T})$ be the result set of the candidate question. We retain it only if the target answer is included and the result set is not overly broad, i.e., $y\in\mathcal{Y}^{(n)}$ and $|\mathcal{Y}^{(n)}|\le K_{\text{ans}}$ for a maximum answer-set size $K_{\text{ans}}$.
\textbf{(b) Non-redundancy.} Let $E_q$ be the topic entities of the question. We extract a shortest supporting path for each $e\in E_q$ and discard the question if a proper subset yields an equally specific answer set:
\begin{equation}
\small
\begin{aligned}
\mathcal{F}_{\mathrm{red}}(Q^{(n)})
= \mathbb{I}\!\Big[
&\exists Z \subsetneq E_q:
\mathcal{Y}^{(n)} \subseteq \mathcal{Y}^{(n)}_{Z}, \\
&|\mathcal{Y}^{(n)}_{Z}| \le |\mathcal{Y}^{(n)}|
\Big],
\end{aligned}
\end{equation}
where $\mathcal{Y}^{(n)}_{Z}$ is the answer set of the query built only from the paths of $Z$ (Appendix~\ref{app:redcheck}).

The triples accumulated for a final question $q$ are its gold supporting triples $\mathcal{S}^{*}_{q}$, and we report the distribution of query-graph types~\citep{GrailQA++} in Appendix~\ref{app:iso_dist}.

\subsection{KG Interaction Agent}
\label{sec:agent}

To make the evidence retrieved for an answer explicit, we develop a KG interaction agent that solves each question through a multi-turn think--act--observe loop with the action space
\begin{equation}
    \mathcal{A}
    =
    \{\texttt{think},\ \texttt{kg-query},\ \texttt{answer}\},
\end{equation}
where \texttt{think} performs internal reasoning, \texttt{kg-query} interacts with the KG, and \texttt{answer} emits the final prediction. The \texttt{kg-query} action is instantiated with seven atomic operations that probe the relations of an entity, expand to its neighboring entities, filter an entity set, and combine entity sets (Appendix~\ref{app:kg_ops}). Every trajectory thus records the returned triples $\mathcal{T}(\tau)$, the only evidence that FoG rewards and evaluates.

\subsection{Evidence-Coverage Process Reward}
\label{sec:training}

To train the agent to retrieve this evidence, we propose a training scheme with outcome and step-validity rewards, the evidence-coverage process reward, and turn-level policy optimization.

\paragraph{Outcome and step-validity rewards.}
We define the outcome reward of a trajectory as
\begin{equation}
\label{eq:rout}
    R_{\text{out}}(\tau)
    =
    \lambda_{\text{f1}}
    \cdot
    \operatorname{F1}(\hat{y},y_q^{*})
    +
    \lambda_{\text{ret}}
    \cdot
    \mathbb{I}[y_q^{*}\in\mathcal{E}(\tau)],
\end{equation}
where the first term rewards a correct answer and the second a reached answer. Neither term checks whether the answer is supported: a \emph{shortcut} trajectory, which outputs and even reaches the gold answer while leaving some of the supporting triples $\mathcal{S}^{*}_{q}$ unretrieved, still earns the full outcome reward.

To stabilize graph interaction, each turn also earns a step-validity reward $r^{\text{step}}_{t}=\lambda_{\text{step}}\,v_t$, where $v_t=1$ if action $a_t$ follows the required format and executes successfully, and $v_t=0$ otherwise. The trajectory-level term is the average $R_{\text{step}}(\tau)=\frac{1}{T}\sum_{t=1}^{T} r^{\text{step}}_{t}$.

\paragraph{Evidence-coverage process reward.}
Unlike the terms above, which treat the KG only as a source of answers, the \textbf{E}vidence-\textbf{C}overage \textbf{P}rocess \textbf{R}eward (ECPR) scores the trajectory against the gold supporting triples $\mathcal{S}^{*}_{q}$ provided by synthesis. We define the evidence coverage of a trajectory as
\begin{equation}
\label{eq:cov}
    \operatorname{Cov}(\tau,\mathcal{S}^{*}_{q})
    =
    \frac{
    |\mathcal{T}(\tau)\cap \mathcal{S}^{*}_{q}|
    }{
    |\mathcal{S}^{*}_{q}|
    },
\end{equation}
where $\mathcal{T}(\tau)$ are all triples retrieved along $\tau$, so a triple that the agent recalls from memory without retrieving it earns no reward. We call $\tau$ evidence-covered if $\operatorname{Cov}(\tau,\mathcal{S}^{*}_{q})=1$, and set $R_{\text{ecpr}}(\tau)=\lambda_{\text{ecpr}}\cdot\operatorname{Cov}(\tau,\mathcal{S}^{*}_{q})$. For dense credit assignment, first-match attribution assigns each gold triple to the step that first retrieves it:
\begin{equation}
    r^{\text{ecpr}}_{t}
    =
    \lambda_{\text{ecpr}}
    \cdot
    \frac{
    \bigl|(\mathcal{T}_t\cap \mathcal{S}^{*}_{q})\setminus \bigcup_{i<t}\mathcal{T}_i\bigr|
    }{
    |\mathcal{S}^{*}_{q}|
    },
\end{equation}
where $\sum_{t} r^{\text{ecpr}}_{t}=R_{\text{ecpr}}(\tau)$, so repeated retrieval of the same triple earns no additional reward. The composite reward of a trajectory is $R(\tau)=R_{\text{out}}(\tau)+R_{\text{step}}(\tau)+R_{\text{ecpr}}(\tau)$. We set $\lambda_{\text{f1}}=0.7$, $\lambda_{\text{ret}}=0.3$, $\lambda_{\text{step}}=0.5$, and $\lambda_{\text{ecpr}}=1$.

\noindent\textbf{Proposition 2.} \textit{ECPR is maximized only by supported answers and rewards each triple once.}\vspace{-2mm}
\begin{proof}\renewcommand{\qedsymbol}{}
We provide experimental results in Sections~\ref{sec:rq2} and~\ref{sec:rq5} and proofs in Appendix~\ref{app:proof2}.
\end{proof}

\paragraph{Turn-level policy optimization.}
Following prior RL-trained KG agents~\citep{Plan-Then-retrieve,KnowCoder-A1,KBQA-R1}, the policy $\pi_{\theta}$ is first initialized by supervised fine-tuning (SFT) on demonstration trajectories that a teacher model produces for the synthesized and training questions, minimizing the negative log-likelihood of the teacher actions. We then optimize it with Group Relative Policy Optimization (GRPO)~\citep{GRPO} and, following KG-R1~\citep{KG-R1}, compute the advantage per turn. For each question $q$, the policy samples $N$ trajectories $\{\tau_i\}_{i=1}^{N}$, and turn $t$ of $\tau_i$ receives the return
\begin{equation}
\label{eq:turn_return}
    G_{i,t}
    =
    R_{\text{out}}(\tau_i)
    + r^{\text{step}}_{i,t}
    + r^{\text{ecpr}}_{i,t},
\end{equation}
i.e., the shared outcome reward plus the process rewards earned by this turn. The advantage $\hat{A}_{i,t}=(G_{i,t}-\mu_q)/(\sigma_q+\delta)$ is normalized with the mean $\mu_q$ and standard deviation $\sigma_q$ of the returns over all turns of the $N$ trajectories. A turn that first retrieves gold evidence therefore receives a higher advantage than the other turns. The policy then maximizes the clipped GRPO objective with these advantages, regularized toward the SFT model (Appendix~\ref{app:advantage}).

\section{Experiments}

This section presents the experimental setup, main results, and analysis. We answer the following research questions (RQs):
\textbf{RQ1:} Does FoG outperform other methods in answer accuracy?
\textbf{RQ2:} Does each component of FoG work?
\textbf{RQ3:} Does evidence-grounded synthesis provide valid evidence and better training data?
\textbf{RQ4:} Does the KG interaction agent retrieve the evidence behind its answers?
\textbf{RQ5:} How does the design of ECPR affect evidence coverage?
\textbf{RQ6:} Does FoG transfer to a KG unseen during training?

\begin{table*}[t]
    \centering
    \footnotesize
    \setlength{\tabcolsep}{6pt}
    \renewcommand{\arraystretch}{1.0}
    \resizebox{\textwidth}{!}{
    \begin{tabular}{l l l cc cc cc}
    \toprule
    \multirow{2}{*}{\textbf{Category}} 
    & \multirow{2}{*}{\textbf{Methods}} 
    & \multirow{2}{*}{\textbf{Backbone}}
    & \multicolumn{2}{c}{\textbf{WikiWebQuestions}}
    & \multicolumn{2}{c}{\textbf{QALD10-en}}
    & \multicolumn{2}{c}{\textbf{GTSQA}} \\
    \cmidrule(lr){4-5} \cmidrule(lr){6-7} \cmidrule(l){8-9}
    & & & \textbf{F1} & \textbf{Hit@1} & \textbf{F1} & \textbf{Hit@1} & \textbf{F1} & \textbf{Hit@1}\\
    \midrule
    
    \multirow{7}{*}{LLM Reasoning}
    & IO Prompt~\cite{qwen} & Qwen2.5-7B-Instruct
    & 28.24	&44.43	&11.05	&12.30	&9.62	&10.23\\
    & IO Prompt~\cite{gpt-4} & GPT-4o-mini
    & 40.31 & 68.10 & 15.34 & 18.78 & 15.37 & 16.46 \\
    & CoT~\cite{CoT} & GPT-4o-mini
    & 41.66 & 58.42 & 22.68 & 26.81 & 18.87 & 20.12 \\
    & Self-Consistency~\cite{Self-Consistency} & GPT-4o-mini
    & 46.34 & 65.20 & 24.27 & 28.57 & 19.07 & 20.21 \\
    & IO Prompt~\cite{deepseekv3} & DeepSeek-V3
    & 42.36 & 59.89 & 18.97 & 21.77 & 24.77 & 26.20 \\
    & CoT~\cite{CoT} & DeepSeek-V3
    & 45.91 &64.60 & 19.02 & 25.28 & 27.66 & 29.41 \\
    & Self-Consistency~\cite{Self-Consistency} & DeepSeek-V3
    & 49.72	&66.79 & 26.25 & 29.97 & 33.59 & 35.82 \\
    \midrule
    
    \multirow{3}{*}{Retrieval-based}
    & RoG~\cite{RoG} & Qwen2.5-7B-Instruct
    & 53.27 & 68.71 & \na & \na & 41.10 & 42.48 \\
    & SubGraphRAG-100~\cite{SubGraphRAG} & Qwen2.5-7B-Instruct
    & 56.84 & 72.36 & \na & \na & 44.27 & 45.93 \\
    & GCR~\cite{GCR} & GPT-4o-mini
    & 55.33 & 71.27 & \na & \na & 32.61 & 37.48 \\
    \midrule
    
    \multirow{5}{*}{Agent-based}
    & ToG~\cite{ToG} & GPT-4o-mini
    & 58.55	&72.75	&30.63	&32.18	&41.08	&42.36 \\
    & PoG~\cite{PoG} & GPT-4o-mini
    & 57.32 & 71.89 & 30.68 & 34.07 & 37.77 & 39.33 \\
    & KG-R1~\cite{KG-R1} & Qwen3-4B-Instruct
    & 64.49  &73.01  &44.07	&45.41	&46.14 &47.73\\
    \rowcolor{lightblue}
    & \textbf{FoG-3B} & Qwen2.5-3B-Instruct
    & \second{66.28} & \second{75.43}& \second{45.63} & \second{48.92} & \second{48.25} & \second{49.20}\\
    \rowcolor{lightblue}
    & \textbf{FoG-4B} & Qwen3-4B-Instruct
    & \best{66.41} &\best{76.93} &\best{47.80} & \best{49.01} &\best{48.66} &\best{50.52} \\
    \bottomrule
    \end{tabular}
    }
    \caption{\textbf{Main results on three KGQA benchmarks.} 
    Best results are shown in \textbf{bold}, and second-best results are \underline{underlined}. Our methods are highlighted in blue. }
    \label{tab:main_result}
    \end{table*}
\begin{table*}[t]
\centering
\renewcommand{\arraystretch}{1.1}
\setlength{\tabcolsep}{6pt}
\resizebox{\textwidth}{!}{
\begin{tabular}{l|cccc|cccc}
    \toprule
    \multirow{2}{*}{Variants}
    & \multicolumn{4}{c|}{WikiWebQuestions}
    & \multicolumn{4}{c}{GTSQA} \\
    \cmidrule(l){2-9}
    & F1 & Hit@1 & RetQ & Cov & F1 & Hit@1 & RetQ & Cov \\
    \midrule
    SFT only
    & 59.95\sd{0.18} & 70.10\sd{0.24} & 79.42\sd{0.31} & 70.63\sd{0.29}
    & 44.73\sd{0.22} & 46.19\sd{0.27} & 58.47\sd{0.33} & 49.85\sd{0.30} \\
    SFT + RL w/o $R_{\text{ecpr}}$
    & 63.87\sd{0.44} & 74.35\sd{0.53} & 83.17\sd{0.49} & 72.91\sd{0.56}
    & 46.02\sd{0.49} & 46.97\sd{0.56} & 61.02\sd{0.52} & 51.43\sd{0.58} \\
    SFT + RL w/o $R_{\text{step}}$
    & 65.02\sd{0.33} & 76.48\sd{0.40} & 85.62\sd{0.38} & 77.25\sd{0.42}
    & 47.96\sd{0.39} & 48.17\sd{0.47} & 63.18\sd{0.44} & 57.21\sd{0.46} \\
    SFT + RL, $R_{\text{ecpr}}\!\rightarrow$ LLM-as-judge
    & 64.95\sd{0.39} & 75.41\sd{0.51} & 84.03\sd{0.45} & 74.12\sd{0.50}
    & 46.88\sd{0.43} & 48.25\sd{0.52} & 62.24\sd{0.48} & 53.36\sd{0.51} \\
    \rowcolor{lightblue}
    \textbf{SFT + RL (full FoG)}
    & \best{66.41}\sd{0.27} & \best{76.93}\sd{0.38} & \best{86.35}\sd{0.35} & \best{78.34}\sd{0.37}
    & \best{48.66}\sd{0.34} & \best{50.52}\sd{0.41} & \best{65.34}\sd{0.39} & \best{58.64}\sd{0.40} \\
    \bottomrule
\end{tabular}}
\caption{\textbf{Ablation of training stages and reward terms.} The LLM-as-judge variant replaces $R_{\text{ecpr}}$ with a model-based process score. Mean and std.\ over three seeds.}
\label{tab:rq4_ecpr_ablation}
\vspace{-5mm}
\end{table*}

\subsection{Experimental Setup}

\paragraph{Datasets.}
We evaluate on three Wikidata-based KGQA benchmarks, WikiWebQuestions~\citep{wikiwebQuestions}, QALD10-en~\citep{qald10en}, and GTSQA~\citep{GTSQA}, and use WebQSP~\citep{webqsp} over Freebase for cross-KG transfer. The synthetic questions do not overlap with the test sets (Appendices~\ref{app:leakage} and~\ref{app:datasets}).

\paragraph{Baselines.}
We compare FoG against \textbf{LLM reasoning} (IO Prompt, CoT~\citep{CoT}, Self-Consistency~\citep{Self-Consistency}), \textbf{retrieval-based} (RoG~\citep{RoG}, SubGraphRAG~\citep{SubGraphRAG}, GCR~\citep{GCR}), and \textbf{agent-based} (ToG~\citep{ToG}, PoG~\citep{PoG}, KG-R1~\citep{KG-R1}) methods. Retrieval-based methods require dataset-specific fine-tuning and are omitted from QALD10-en, whose training split has only 387 examples (Appendix~\ref{app:baselines}).

\paragraph{Evaluation Metrics.}
We report \textbf{Hit@1} and token-level \textbf{F1} for answer accuracy. For evidence coverage, we report \textbf{Cov}, the fraction of gold supporting triples retrieved along the trajectory, and \textbf{RetQ}, the fraction of questions whose gold answer is returned by a KG query. GTSQA provides gold supporting triples natively, and we construct them for WikiWebQuestions with an LLM verifier that agrees with human annotators on 93.0\% of 200 audited paths (Appendices~\ref{app:metrics}, \ref{app:gold_supporting_triples}, and~\ref{app:human_audit}).

\paragraph{Implementation Details.}
We train FoG on Qwen2.5-3B-Instruct and Qwen3-4B-Instruct with LoRA-based SFT followed by GRPO, sampling $N{=}4$ trajectories of at most 10 KG turns per question for 450 steps. All methods are evaluated with a \emph{single inference run} per question (Appendix~\ref{app:training}).

\subsection{Main Results (RQ1)}
\label{sec:rq1}

As shown in Table~\ref{tab:main_result}, we compare FoG with all baselines in answer accuracy.

\paragraph{Comparison across methods.}
FoG achieves the best F1 and Hit@1 on all three benchmarks, exceeding the strongest baseline KG-R1 by up to +3.73 F1 and +3.92 Hit@1. As KG-R1 is also RL-trained on the same Qwen3-4B backbone, this indicates that training an agent to retrieve evidence also makes its answers more accurate.

\paragraph{Comparison across backbones.}
FoG-3B still exceeds the 4B KG-R1 on all three benchmarks, and FoG-4B surpasses GPT-4o-mini agents by at least +7.58 F1 and 7B retrieval-based methods by at least +4.39 F1. This suggests that how an agent is trained to retrieve matters more than its backbone size.

\subsection{Ablation Study (RQ2)}
\label{sec:rq2}
\begin{figure*}[t]
    \centering
    \includegraphics[width=0.32\textwidth]{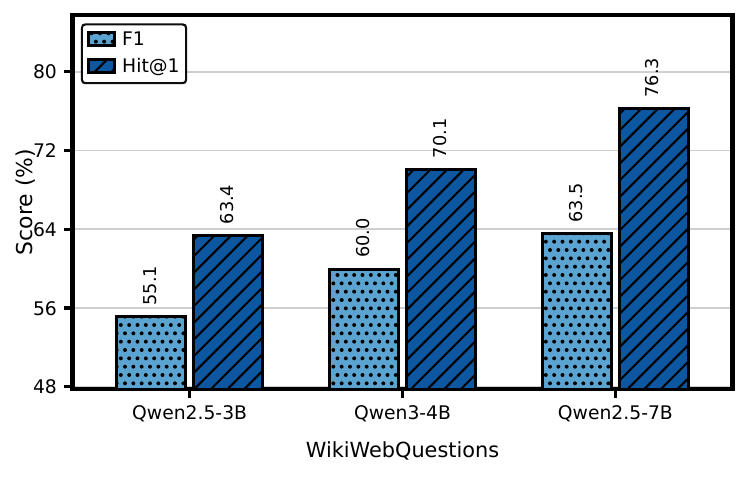}\hfill
    \includegraphics[width=0.32\textwidth]{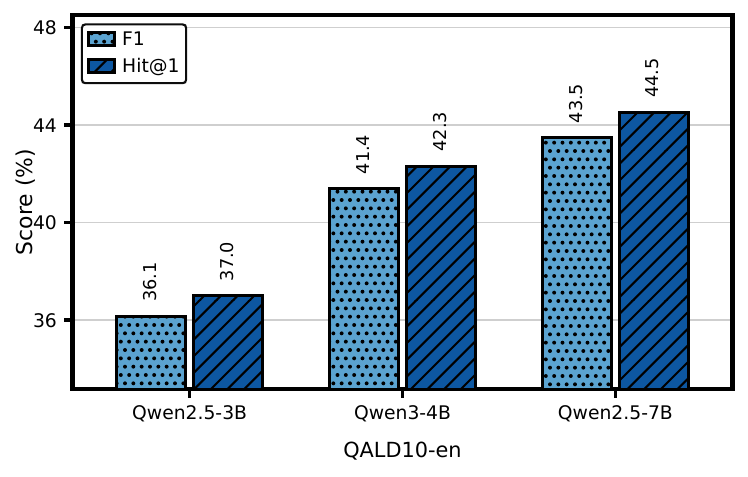}\hfill
    \includegraphics[width=0.32\textwidth]{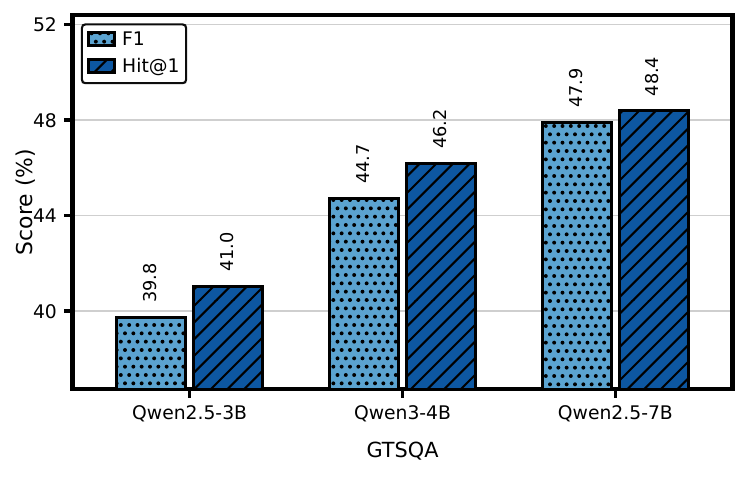}
    \caption{\textbf{Effect of model scale.} F1 and Hit@1 of backbones of different sizes after SFT on the same data.}
    \label{fig:model_scale}
    \vspace{-5mm}
\end{figure*}

As shown in Table~\ref{tab:rq4_ecpr_ablation}, we ablate the training stages and reward terms on the two benchmarks with gold supporting triples, under the same backbone, SFT checkpoint, RL data, and rollout budget, so the variants differ only in the reward.

\paragraph{Impact of RL.}
Full FoG improves over SFT only by +6.83 and +4.33 Hit@1, but RL without $R_{\text{ecpr}}$ raises Cov by only +2.28 and +1.58. This is because the outcome reward is paid in full once the answer is reached, whatever evidence is retrieved.

\paragraph{Impact of $R_{\text{ecpr}}$.}
Removing $R_{\text{ecpr}}$ causes the largest drop among the reward terms, and Cov drops more (5.43 and 7.21) than Hit@1 (2.58 and 3.55). As both variants share the data and the SFT checkpoint, this confirms that the coverage gain comes from the reward itself.

\paragraph{Impact of $R_{\text{step}}$.}
Removing $R_{\text{step}}$ lowers Hit@1 by 0.45 and 2.35 but Cov by only 1.09 and 1.43, indicating that step validity keeps the interaction executable but does not drive coverage.

\paragraph{ECPR vs.\ LLM-as-judge.}
An LLM-as-judge score of the reasoning process~\citep{KG-Hopper} raises Cov by only 1.21 and 1.93 over no process supervision and stays 4.22 and 5.28 below ECPR. This suggests that judging how plausible the reasoning reads is a weaker signal than ECPR.

\subsection{Evidence-Grounded Synthesis (RQ3)}
\label{sec:rq3}

As shown in Table~\ref{tab:rq3_synth_data} and Figure~\ref{fig:model_scale}, we fix a budget of 6,588 SFT questions and vary the data source (Train-Only, Synth-Only, or a mix of both), and apply the same SFT data to three backbone scales.

\paragraph{Validity of embedded evidence.}
The supporting triples of every retained question reproduce its answer under SPARQL execution, and the redundancy check misses a shortcut in only 1.5\% of 200 audited questions. This confirms that the embedded triples are a valid reward target without annotation.

\begin{table}[t]
\centering
\footnotesize
\setlength{\tabcolsep}{6pt}
\renewcommand{\arraystretch}{1.1}
\resizebox{\columnwidth}{!}{
\begin{tabular}{l|cc|cc|cc}
    \toprule
    \multirow{2}{*}{Data Source}
    & \multicolumn{2}{c|}{WikiWebQuestions}
    & \multicolumn{2}{c|}{QALD10-en}
    & \multicolumn{2}{c}{GTSQA} \\
    \cmidrule(l){2-7}
   & F1 & Hit@1 & F1 & Hit@1 & F1 & Hit@1 \\
    \midrule
    Train-Only
    & 53.58 & 64.51 & 36.94 & 37.86 & 41.33 & 41.52 \\
    Synth-Only
    & 57.84 & 68.17 & 39.52 & 40.36 & 43.21 & 44.58 \\
    \rowcolor{lightblue}
    \textbf{Mixed}
    & \best{59.95} & \best{70.10} & \best{41.38} & \best{42.31} & \best{44.73} & \best{46.19} \\
    \bottomrule
\end{tabular}}
\caption{\textbf{Impact of synthetic training data.}}
\label{tab:rq3_synth_data}
\end{table}

\begin{table}[!t]
\centering
\renewcommand{\arraystretch}{0.9}
\resizebox{\columnwidth}{!}{          
\begin{tabular}{l|ccc|ccc}
    \toprule
    \multirow{2}{*}{Methods} 
    & \multicolumn{3}{c|}{WikiWebQuestions} 
    & \multicolumn{3}{c}{GTSQA} \\ 
    \cmidrule(l){2-7} 
    & Hit@1 & RetQ & Cov
    & Hit@1 & RetQ & Cov \\ 
    \midrule
    PoG                        
    & 71.89 & 75.66 & 69.85 & 39.33 & 35.70 & 25.38 \\
    ToG                        
    & 72.75 & 73.53 & 68.79 & 42.36 & 50.99 & 37.34 \\
    KG-R1
    & 73.01 & 82.28 & 74.50 & 47.73 & 61.41 & 54.05 \\
        \rowcolor{lightblue}
    FoG-4B
    & \textbf{76.93} & \textbf{86.35} & \textbf{78.34} & \textbf{50.52} & \textbf{65.34} & \textbf{58.64} \\
    \bottomrule
\end{tabular}
}
\caption{\textbf{Answer accuracy and evidence coverage of agent-based methods.}}
\label{tab:rq2_faithful}
\end{table}

\paragraph{Synthetic vs.\ original data.}
Synth-Only outperforms Train-Only by +3.66, +2.50, and +3.06 Hit@1 with no in-domain question, indicating that questions built by backward expansion transfer to human-written benchmarks. Mixed performs best, so the two sources are complementary (Appendix~\ref{app:synth_cost}).

\paragraph{Performance across model scales.}
Hit@1 improves steadily from 3B to 7B (+12.89, +7.52, and +7.36), indicating that the synthetic data is not backbone-specific.

\subsection{Analysis of KG Interaction Agent (RQ4)}
\label{sec:rq4}

\begin{figure*}[t]
    \centering
    \includegraphics[width=\linewidth]{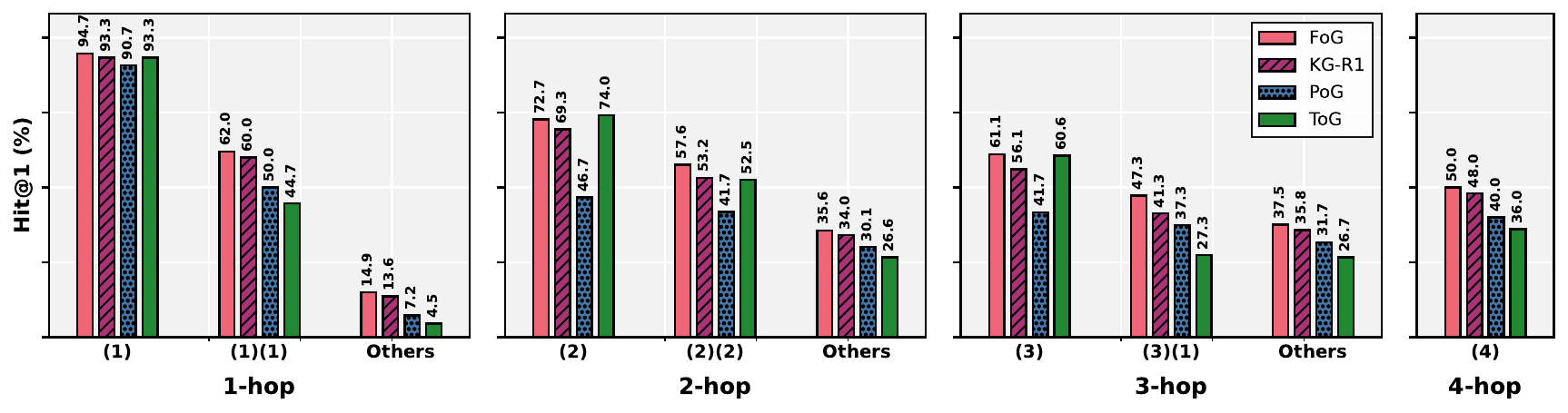}
    \caption{\textbf{Performance under structural complexity on GTSQA} by reasoning hop and isomorphism type.}
    \label{fig:complexity}
    \vspace{-5mm}
\end{figure*}

As shown in Table~\ref{tab:rq2_faithful} and Figure~\ref{fig:complexity}, we examine how much evidence agent-based methods retrieve.

\paragraph{Answers without evidence.}
On GTSQA, PoG attains 39.33 Hit@1 but reaches the gold answer on only 35.70\% of questions, so at least 9.2\% of its correct answers never appear in any KG observation, and the RL-trained KG-R1 still leaves 45.95\% of the supporting triples unretrieved. Neither prompting nor outcome rewards thus make an agent use the KG as a source of evidence.

\paragraph{Evidence coverage.}
FoG attains the highest Cov and RetQ on both benchmarks (+3.84 and +4.59 Cov over KG-R1) together with the highest Hit@1, indicating that evidence coverage does not come at the cost of answer accuracy (Appendix~\ref{app:casestudy}).

\paragraph{Performance under structural complexity.}
FoG leads KG-R1 on all 10 groups and is the best method on 9. Its margin over KG-R1 grows from +1.4 Hit@1 on $(1)$ to +5.0 on $(3)$ and +6.0 on the more branched $(3)(1)$, questions that depend on more supporting triples. This suggests that the more evidence an answer requires, the more the agent benefits from rewards for retrieving it, although the margin narrows to +2.0 on $(4)$.

\begin{figure*}[t]
    \centering
    \includegraphics[width=0.245\textwidth]{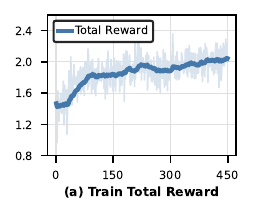}\hfill
    \includegraphics[width=0.245\textwidth]{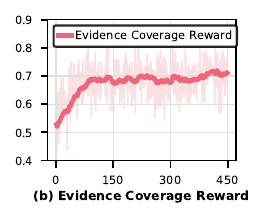}\hfill
    \includegraphics[width=0.245\textwidth]{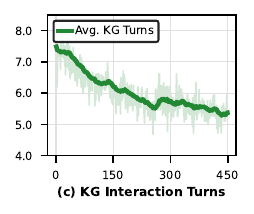}\hfill
    \includegraphics[width=0.245\textwidth]{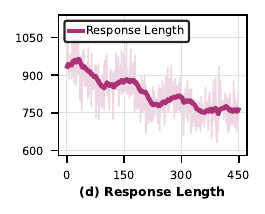}
    \caption{\textbf{Training dynamics of FoG.} Rewards rise while interaction turns and response length decline.}
    \label{fig:training_dynamics}
    \vspace{-5mm}
\end{figure*}

\subsection{Analysis of ECPR (RQ5)}
\label{sec:rq5}
\begin{table}[t]
\centering
\small
\begin{tabular}{lccc}
\toprule
\textbf{Credit assignment} & \textbf{Hit@1} & \textbf{Cov} & \textbf{Dup.\ rate} \\
\midrule
Trajectory-level & 46.69 & 54.29 & 11.8\% \\
Every-match      & 47.01 & 52.21 & 29.4\% \\
\rowcolor{lightblue}
\textbf{First-match (ours)} & 50.52 & 58.64 & 8.6\% \\
\bottomrule
\end{tabular}
\caption{\textbf{Credit-assignment strategies for ECPR on GTSQA.} Dup.\ rate is the fraction of KG queries that re-retrieve an already retrieved gold triple.}
\label{tab:credit_assign}
\vspace{-5mm}
\end{table}

As shown in Table~\ref{tab:credit_assign} and Figure~\ref{fig:training_dynamics}, we compare first-match with \emph{trajectory-level} attribution, which credits coverage to the final turn, and \emph{every-match} attribution, which rewards every retrieval.

\paragraph{Comparison across attribution strategies.}
First-match achieves the best Hit@1 and Cov. Trajectory-level attribution loses 4.35 Cov, as turns that retrieve evidence earn the same return as the others. Every-match raises the duplicate-retrieval rate from 8.6\% to 29.4\% and falls below trajectory-level in Cov, as it rewards re-querying known evidence. This indicates that dense credit must pay each triple once, or it rewards redundant retrieval rather than coverage.

\paragraph{Training dynamics.}
Rewards rise during RL while KG turns and response length decline, indicating that the agent learns a precise retrieval policy that covers more evidence with fewer queries.

\begin{table}[t]
\centering
\footnotesize
\setlength{\tabcolsep}{5pt}
\renewcommand{\arraystretch}{1.1}
\resizebox{\columnwidth}{!}{
\begin{tabular}{l l l c c}
    \toprule
    \multirow{2}{*}{\textbf{Method}}
    & \multirow{2}{*}{\textbf{Train KG}}
    & \multirow{2}{*}{\textbf{Backbone}}
    & \textbf{WebQSP} & \textbf{WebQSP} \\
    & & & \textbf{train} & \textbf{Hit@1} \\
    \midrule
    ToG~\cite{ToG}          & None     & GPT-4                & No  & 82.6 \\
    RoG~\cite{RoG}          & Freebase & LLaMA2-Chat-7B       & Yes & 85.7 \\
    KG-R1~\cite{KG-R1}      & Freebase & Qwen2.5-3B-Instruct  & Yes & 82.8 \\
    \midrule
    FoG SFT-only            & Wikidata & Qwen3-4B-Instruct    & No  & 71.6 \\
    \textbf{FoG full}       & Wikidata & Qwen3-4B-Instruct    & No  & 77.2 \\
    \bottomrule
\end{tabular}}
\caption{\textbf{Zero-shot cross-KG transfer to Freebase (WebQSP).} Methods developed on Freebase are reference points, not matched baselines.}
\label{tab:rq5_crosskg}
\vspace{-5mm}
\end{table}

\subsection{Cross-KG Transfer (RQ6)}
\label{sec:rq6}

As shown in Table~\ref{tab:rq5_crosskg}, we evaluate the frozen FoG zero-shot on WebQSP over a Freebase backend.

\paragraph{Zero-shot transfer.}
FoG reaches 77.2 Hit@1 on an unseen KG, close to the methods developed on Freebase (82.6--85.7), and RL with ECPR contributes +5.6 Hit@1 over SFT only. This suggests that ECPR teaches the habit of retrieving evidence before answering, not the relations of one schema.

\section{Conclusion}

In this work, we present FoG, a framework that trains KG agents to retrieve the evidence behind their answers. By introducing evidence-grounded synthesis, a KG interaction agent, and an evidence-coverage process reward, FoG uses the KG as a source of evidence rather than only as a source of answers. Experiments on three KGQA benchmarks demonstrate consistent improvements in answer accuracy, evidence coverage, and cross-KG transfer.

\section*{Limitations}

\paragraph{KG completeness and freshness.}
FoG uses Wikidata triples as verifiable references for evidence-covered supervision, but real-world knowledge graphs may be incomplete, outdated, or noisy. When the required evidence is absent or incorrect in the graph, the agent may still fail despite following the intended evidence path. A natural extension is to equip the agent with a mechanism for detecting insufficient on-graph evidence and invoking external retrieval when needed.

\paragraph{Dependence on teacher models.}
Like other KG agents distilled from teacher trajectories~\citep{KnowCoder-A1,SoG-R1}, FoG currently relies on an external teacher LLM (Kimi-K2) for question synthesis and trajectory distillation. Although symbolic checks filter invalid questions and evidence paths, the synthesis process still depends on the generation quality of external teachers. One promising direction is to replace teacher-driven synthesis with self-play, where agents generate, explore, and verify candidate questions through graph execution and symbolic filtering.

\paragraph{Long-horizon graph interaction.}
FoG caps graph interactions at 10 turns, which is sufficient for the 1--4-hop questions studied in our benchmarks. More open-ended KG reasoning may require longer evidence chains and broader graph exploration. Extending FoG to such settings requires balancing search depth against error accumulation, noisy observations, and sparser rewards.

\section*{Ethical Considerations}

All experiments use publicly available KGQA datasets derived from Wikidata (CC0), QALD10-en (MIT License), and WikiWebQuestions, GTSQA, and WebQSP (publicly available for research). Backbone models Qwen2.5 and Qwen3 are released under Apache~2.0; GPT-4o-mini, DeepSeek-V3, and the teacher Kimi-K2 are accessed via official APIs under their terms of service. Our use of all existing artifacts is consistent with their intended research purposes. The datasets contain only public encyclopedic facts from structured KG triples, with no private or user-generated data. The only human involvement is the audit of constructed evidence paths (Appendix~\ref{app:human_audit}), in which three graduate annotators judged whether a KG path supports its question. The annotators took part voluntarily, the task exposed them to no sensitive content, and no personal data was collected. This work involves no crowdsourcing and no user study. FoG is intended to make the answers of KG agents traceable to retrieved evidence. Since the KG may be outdated or incorrect, this evidence supports human verification rather than replaces it. We foresee no dual-use concerns.

\bibliography{refs}

\begin{thebibliography}{54}
\providecommand{\natexlab}[1]{#1}

\bibitem[{Agrawal et~al.(2024)Agrawal, Kumarage, Alghamdi, and Liu}]{agrawal2024mindful}
Garima Agrawal, Tharindu Kumarage, Zeyad Alghamdi, and Huan Liu. 2024.
\newblock \href {https://doi.org/10.1109/FLLM63129.2024.10852457} {Mindful-rag: {A} study of points of failure in retrieval augmented generation}.
\newblock In \emph{2nd International Conference on Foundation and Large Language Models, {FLLM} 2024, Dubai, United Arab Emirates, November 26-29, 2024}, pages 607--611. {IEEE}.

\bibitem[{Alansari and Luqman(2026)}]{hallucination_2}
Aisha Alansari and Hamzah Luqman. 2026.
\newblock \href {https://doi.org/10.1016/J.COSREV.2026.100970} {Large language models hallucination: {A} comprehensive survey}.
\newblock \emph{Comput. Sci. Rev.}, 61:100970.

\bibitem[{Cabello et~al.(2024)Cabello, Martin{-}Turrero, Akujuobi, S{\o}gaard, and Bobed}]{cabello2024meg}
Laura Cabello, Carmen Martin{-}Turrero, Uchenna Akujuobi, Anders S{\o}gaard, and Carlos Bobed. 2024.
\newblock \href {https://doi.org/10.48550/ARXIV.2411.03883} {{MEG:} medical knowledge-augmented large language models for question answering}.
\newblock \emph{CoRR}, abs/2411.03883.

\bibitem[{Cattaneo et~al.(2026)Cattaneo, Luschi, and Justus}]{GTSQA}
Alberto Cattaneo, Carlo Luschi, and Daniel Justus. 2026.
\newblock \href {https://openreview.net/forum?id=n0Nh5SeR3i} {Ground-truth subgraphs for better training and evaluation of knowledge graph augmented llms}.
\newblock \emph{Trans. Mach. Learn. Res.}, 2026.

\bibitem[{Chen et~al.(2024)Chen, Tong, Jin, Sun, Ye, and Xiong}]{PoG}
Liyi Chen, Panrong Tong, Zhongming Jin, Ying Sun, Jieping Ye, and Hui Xiong. 2024.
\newblock \href {http://papers.nips.cc/paper\_files/paper/2024/hash/4254e856d01a5e7b7ea050477c3ef9b9-Abstract-Conference.html} {Plan-on-graph: Self-correcting adaptive planning of large language model on knowledge graphs}.
\newblock In \emph{Advances in Neural Information Processing Systems 37: Annual Conference on Neural Information Processing Systems 2024, NeurIPS 2024, Vancouver, BC, Canada, December 10 - 15, 2024}.

\bibitem[{Chen et~al.(2025)Chen, Wang, Li, Zhang, Ding, Yang, Xu, Jin, and Guo}]{KnowCoder-A1}
Zhuo Chen, Fei Wang, Zixuan Li, Zhao Zhang, Weiwei Ding, Chuanguang Yang, YongJun Xu, Xiaolong Jin, and Jiafeng Guo. 2025.
\newblock \href {https://doi.org/10.48550/ARXIV.2510.25101} {Knowcoder-a1: Incentivizing agentic reasoning capability with outcome supervision for {KBQA}}.
\newblock \emph{CoRR}, abs/2510.25101.

\bibitem[{DeepSeek{-}AI(2024)}]{deepseekv3}
DeepSeek{-}AI. 2024.
\newblock \href {https://doi.org/10.48550/ARXIV.2412.19437} {Deepseek-v3 technical report}.
\newblock \emph{CoRR}, abs/2412.19437.

\bibitem[{Dutt et~al.(2023)Dutt, Khosla, Kumar, and Gangadharaiah}]{GrailQA++}
Ritam Dutt, Sopan Khosla, Vinayshekhar~Bannihatti Kumar, and Rashmi Gangadharaiah. 2023.
\newblock \href {https://doi.org/10.18653/V1/2023.IJCNLP-MAIN.58} {Grailqa++: {A} challenging zero-shot benchmark for knowledge base question answering}.
\newblock In \emph{Proceedings of the 13th International Joint Conference on Natural Language Processing and the 3rd Conference of the Asia-Pacific Chapter of the Association for Computational Linguistics, {IJCNLP} 2023 -Volume 1: Long Papers, Nusa Dua, Bali, November 1 - 4, 2023}, pages 897--909. Association for Computational Linguistics.

\bibitem[{GLM(2025)}]{GLM}
GLM. 2025.
\newblock \href {https://doi.org/10.48550/ARXIV.2508.06471} {{GLM-4.5:} agentic, reasoning, and coding {(ARC)} foundation models}.
\newblock \emph{CoRR}, abs/2508.06471.

\bibitem[{Gu et~al.(2021)Gu, Kase, Vanni, Sadler, Liang, Yan, and Su}]{grailqa}
Yu~Gu, Sue Kase, Michelle Vanni, Brian Sadler, Percy Liang, Xifeng Yan, and Yu~Su. 2021.
\newblock Beyond iid: three levels of generalization for question answering on knowledge bases.
\newblock In \emph{Proceedings of the web conference 2021}, pages 3477--3488.

\bibitem[{Han et~al.(2026)Han, Guo, Shomer, Wang, Chu, Li, Ma, and Tang}]{RoE}
Haoyu Han, Kai Guo, Harry Shomer, Yu~Wang, Yucheng Chu, Hang Li, Li~Ma, and Jiliang Tang. 2026.
\newblock \href {https://doi.org/10.1145/3774904.3792684} {Reasoning by exploration: {A} unified approach to retrieval and generation over graphs}.
\newblock In \emph{Proceedings of the {ACM} Web Conference 2026, {WWW} 2026, Dubai, United Arab Emirates, originally scheduled for April 13-17, 2026, rescheduled for June 29 - July 3, 2026}, pages 2408--2418. {ACM}.

\bibitem[{He et~al.(2024)He, Tian, Sun, Chawla, Laurent, LeCun, Bresson, and Hooi}]{G-Retrieval}
Xiaoxin He, Yijun Tian, Yifei Sun, Nitesh~V. Chawla, Thomas Laurent, Yann LeCun, Xavier Bresson, and Bryan Hooi. 2024.
\newblock \href {http://papers.nips.cc/paper\_files/paper/2024/hash/efaf1c9726648c8ba363a5c927440529-Abstract-Conference.html} {G-retriever: Retrieval-augmented generation for textual graph understanding and question answering}.
\newblock In \emph{Advances in Neural Information Processing Systems 37: Annual Conference on Neural Information Processing Systems 2024, NeurIPS 2024, Vancouver, BC, Canada, December 10 - 15, 2024}.

\bibitem[{Hu et~al.(2024)Hu, Gao, Gao, Chen, and Huang}]{hu2024large}
Peng Hu, Changjiang Gao, Ruiqi Gao, Jiajun Chen, and Shujian Huang. 2024.
\newblock \href {https://doi.org/10.18653/V1/2024.FINDINGS-EMNLP.178} {Large language models are limited in out-of-context knowledge reasoning}.
\newblock In \emph{Findings of the Association for Computational Linguistics: {EMNLP} 2024, Miami, Florida, USA, November 12-16, 2024}, volume {EMNLP} 2024 of \emph{Findings of {ACL}}, pages 3144--3155. Association for Computational Linguistics.

\bibitem[{Huang et~al.(2025)Huang, Yu, Ma, Zhong, Feng, Wang, Chen, Peng, Feng, Qin, and Liu}]{hallucination_1}
Lei Huang, Weijiang Yu, Weitao Ma, Weihong Zhong, Zhangyin Feng, Haotian Wang, Qianglong Chen, Weihua Peng, Xiaocheng Feng, Bing Qin, and Ting Liu. 2025.
\newblock \href {https://doi.org/10.1145/3703155} {A survey on hallucination in large language models: Principles, taxonomy, challenges, and open questions}.
\newblock \emph{{ACM} Trans. Inf. Syst.}, 43(2):42:1--42:55.

\bibitem[{Jiang et~al.(2025)Jiang, Zhou, Zhao, Song, Zhu, Zhu, and Wen}]{kg-agent}
Jinhao Jiang, Kun Zhou, Xin Zhao, Yang Song, Chen Zhu, Hengshu Zhu, and Ji{-}Rong Wen. 2025.
\newblock \href {https://doi.org/10.18653/V1/2025.ACL-LONG.468} {Kg-agent: An efficient autonomous agent framework for complex reasoning over knowledge graph}.
\newblock In \emph{Proceedings of the 63rd Annual Meeting of the Association for Computational Linguistics (Volume 1: Long Papers), {ACL} 2025, Vienna, Austria, July 27 - August 1, 2025}, pages 9505--9523. Association for Computational Linguistics.

\bibitem[{Jin et~al.(2024)Jin, Xie, Zhang, Roy, Zhang, Li, Li, Tang, Wang, Meng, and Han}]{GoT}
Bowen Jin, Chulin Xie, Jiawei Zhang, Kashob~Kumar Roy, Yu~Zhang, Zheng Li, Ruirui Li, Xianfeng Tang, Suhang Wang, Yu~Meng, and Jiawei Han. 2024.
\newblock \href {https://doi.org/10.18653/V1/2024.FINDINGS-ACL.11} {Graph chain-of-thought: Augmenting large language models by reasoning on graphs}.
\newblock In \emph{Findings of the Association for Computational Linguistics, {ACL} 2024, Bangkok, Thailand and virtual meeting, August 11-16, 2024}, volume {ACL} 2024 of \emph{Findings of {ACL}}, pages 163--184. Association for Computational Linguistics.

\bibitem[{Jullien et~al.(2025)Jullien, Valentino, and Freitas}]{jullien2025knowledge}
Ma{\"{e}}l Jullien, Marco Valentino, and Andr{\'{e}} Freitas. 2025.
\newblock \href {https://doi.org/10.48550/ARXIV.2508.10777} {The knowledge-reasoning dissociation: Fundamental limitations of llms in clinical natural language inference}.
\newblock \emph{CoRR}, abs/2508.10777.

\bibitem[{Li et~al.(2025)Li, Miao, and Li}]{SubGraphRAG}
Mufei Li, Siqi Miao, and Pan Li. 2025.
\newblock \href {https://openreview.net/forum?id=JvkuZZ04O7} {Simple is effective: The roles of graphs and large language models in knowledge-graph-based retrieval-augmented generation}.
\newblock In \emph{The Thirteenth International Conference on Learning Representations, {ICLR} 2025, Singapore, April 24-28, 2025}. OpenReview.net.

\bibitem[{Lin et~al.(2025)Lin, Liu, Song, Wang, Shun, and Zhu}]{KG-R1}
Junhong Lin, Shicheng Liu, Jinyeop Song, Song Wang, Julian Shun, and Yada Zhu. 2025.
\newblock Efficient and transferable agentic knowledge graph rag via reinforcement learning.
\newblock \emph{arXiv preprint arXiv:2509.26383}.

\bibitem[{Liu et~al.(2026)Liu, Lin, Shang, Li, Wang, and Cao}]{PathMind}
Yu~Liu, Xixun Lin, Yanmin Shang, Yangxi Li, Shi Wang, and Yanan Cao. 2026.
\newblock \href {https://doi.org/10.1609/AAAI.V40I18.38565} {Pathmind: {A} retrieve-prioritize-reason framework for knowledge graph reasoning with large language models}.
\newblock In \emph{Fortieth {AAAI} Conference on Artificial Intelligence, Thirty-Eighth Conference on Innovative Applications of Artificial Intelligence, Sixteenth Symposium on Educational Advances in Artificial Intelligence, {AAAI} 2026, Singapore, January 20-27, 2026}, pages 15386--15393. {AAAI} Press.

\bibitem[{Luo et~al.(2024)Luo, Li, Haffari, and Pan}]{RoG}
Linhao Luo, Yuan{-}Fang Li, Gholamreza Haffari, and Shirui Pan. 2024.
\newblock \href {https://openreview.net/forum?id=ZGNWW7xZ6Q} {Reasoning on graphs: Faithful and interpretable large language model reasoning}.
\newblock In \emph{The Twelfth International Conference on Learning Representations, {ICLR} 2024, Vienna, Austria, May 7-11, 2024}. OpenReview.net.

\bibitem[{Luo et~al.(2025)Luo, Zhao, Haffari, Li, Gong, and Pan}]{GCR}
Linhao Luo, Zicheng Zhao, Gholamreza Haffari, Yuan{-}Fang Li, Chen Gong, and Shirui Pan. 2025.
\newblock \href {https://proceedings.mlr.press/v267/luo25t.html} {Graph-constrained reasoning: Faithful reasoning on knowledge graphs with large language models}.
\newblock In \emph{Forty-second International Conference on Machine Learning, {ICML} 2025, Vancouver, BC, Canada, July 13-19, 2025}, volume 267 of \emph{Proceedings of Machine Learning Research}. {PMLR} / OpenReview.net.

\bibitem[{Mavromatis and Karypis(2025)}]{GNN-RAG}
Costas Mavromatis and George Karypis. 2025.
\newblock \href {https://doi.org/10.18653/V1/2025.FINDINGS-ACL.856} {{GNN-RAG:} graph neural retrieval for efficient large language model reasoning on knowledge graphs}.
\newblock In \emph{Findings of the Association for Computational Linguistics, {ACL} 2025, Vienna, Austria, July 27 - August 1, 2025}, volume {ACL} 2025 of \emph{Findings of {ACL}}, pages 16682--16699. Association for Computational Linguistics.

\bibitem[{Mirzadeh et~al.(2025)Mirzadeh, Alizadeh, Shahrokhi, Tuzel, Bengio, and Farajtabar}]{mirzadeh2025gsm}
Iman Mirzadeh, Keivan Alizadeh, Hooman Shahrokhi, Oncel Tuzel, Samy Bengio, and Mehrdad Farajtabar. 2025.
\newblock \href {https://openreview.net/forum?id=AjXkRZIvjB} {Gsm-symbolic: Understanding the limitations of mathematical reasoning in large language models}.
\newblock In \emph{The Thirteenth International Conference on Learning Representations, {ICLR} 2025, Singapore, April 24-28, 2025}. OpenReview.net.

\bibitem[{Nguyen et~al.(2024)Nguyen, Luo, Shiri, Phung, Li, Vu, and Haffari}]{nguyen2024direct}
Thi Nguyen, Linhao Luo, Fatemeh Shiri, Dinh Phung, Yuan{-}Fang Li, Thuy{-}Trang Vu, and Gholamreza Haffari. 2024.
\newblock \href {https://doi.org/10.18653/V1/2024.FINDINGS-ACL.168} {Direct evaluation of chain-of-thought in multi-hop reasoning with knowledge graphs}.
\newblock In \emph{Findings of the Association for Computational Linguistics, {ACL} 2024, Bangkok, Thailand and virtual meeting, August 11-16, 2024}, volume {ACL} 2024 of \emph{Findings of {ACL}}, pages 2862--2883. Association for Computational Linguistics.

\bibitem[{OpenAI(2023)}]{gpt-4}
OpenAI. 2023.
\newblock \href {https://doi.org/10.48550/ARXIV.2303.08774} {{GPT-4} technical report}.
\newblock \emph{CoRR}, abs/2303.08774.

\bibitem[{Pan et~al.(2024)Pan, Luo, Wang, Chen, Wang, and Wu}]{roadmap}
Shirui Pan, Linhao Luo, Yufei Wang, Chen Chen, Jiapu Wang, and Xindong Wu. 2024.
\newblock \href {https://doi.org/10.1109/TKDE.2024.3352100} {Unifying large language models and knowledge graphs: {A} roadmap}.
\newblock \emph{{IEEE} Trans. Knowl. Data Eng.}, 36(7):3580--3599.

\bibitem[{Petroni et~al.(2021)Petroni, Piktus, Fan, Lewis, Yazdani, Cao, Thorne, Jernite, Karpukhin, Maillard, Plachouras, Rockt{\"{a}}schel, and Riedel}]{petroni2021kilt}
Fabio Petroni, Aleksandra Piktus, Angela Fan, Patrick Lewis, Majid Yazdani, Nicola~De Cao, James Thorne, Yacine Jernite, Vladimir Karpukhin, Jean Maillard, Vassilis Plachouras, Tim Rockt{\"{a}}schel, and Sebastian Riedel. 2021.
\newblock \href {https://doi.org/10.18653/V1/2021.NAACL-MAIN.200} {{KILT:} a benchmark for knowledge intensive language tasks}.
\newblock In \emph{Proceedings of the 2021 Conference of the North American Chapter of the Association for Computational Linguistics: Human Language Technologies, {NAACL-HLT} 2021, Online, June 6-11, 2021}, pages 2523--2544. Association for Computational Linguistics.

\bibitem[{Shao et~al.(2024)Shao, Wang, Zhu, Xu, Song, Zhang, Li, Wu, and Guo}]{GRPO}
Zhihong Shao, Peiyi Wang, Qihao Zhu, Runxin Xu, Junxiao Song, Mingchuan Zhang, Y.~K. Li, Y.~Wu, and Daya Guo. 2024.
\newblock \href {https://doi.org/10.48550/ARXIV.2402.03300} {Deepseekmath: Pushing the limits of mathematical reasoning in open language models}.
\newblock \emph{CoRR}, abs/2402.03300.

\bibitem[{Sheng et~al.(2025)Sheng, Zhang, Ye, Wu, Zhang, Zhang, Peng, Lin, and Wu}]{sheng2025hybridflow}
Guangming Sheng, Chi Zhang, Zilingfeng Ye, Xibin Wu, Wang Zhang, Ru~Zhang, Yanghua Peng, Haibin Lin, and Chuan Wu. 2025.
\newblock \href {https://doi.org/10.1145/3689031.3696075} {Hybridflow: {A} flexible and efficient {RLHF} framework}.
\newblock In \emph{Proceedings of the Twentieth European Conference on Computer Systems, EuroSys 2025, Rotterdam, The Netherlands, 30 March 2025 - 3 April 2025}, pages 1279--1297. {ACM}.

\bibitem[{Song et~al.(2026)Song, Liu, Guti{\'{e}}rrez{-}Basulto, Hu, Xie, Peng, Ananiadou, and Pan}]{Plan-Then-retrieve}
Yanlin Song, Ben Liu, V{\'{\i}}ctor Guti{\'{e}}rrez{-}Basulto, Zhiwei Hu, Qianqian Xie, Min Peng, Sophia Ananiadou, and Jeff~Z. Pan. 2026.
\newblock \href {https://doi.org/10.1145/3774904.3792191} {Plan then retrieve: Reinforcement learning-guided complex reasoning over knowledge graphs}.
\newblock In \emph{Proceedings of the {ACM} Web Conference 2026, {WWW} 2026, Dubai, United Arab Emirates, originally scheduled for April 13-17, 2026, rescheduled for June 29 - July 3, 2026}, pages 3666--3676. {ACM}.

\bibitem[{Sui et~al.(2025)Sui, He, Liu, He, Wang, and Hooi}]{sui2025fidelis}
Yuan Sui, Yufei He, Nian Liu, Xiaoxin He, Kun Wang, and Bryan Hooi. 2025.
\newblock \href {https://doi.org/10.18653/V1/2025.FINDINGS-ACL.436} {Fidelis: Faithful reasoning in large language models for knowledge graph question answering}.
\newblock In \emph{Findings of the Association for Computational Linguistics, {ACL} 2025, Vienna, Austria, July 27 - August 1, 2025}, volume {ACL} 2025 of \emph{Findings of {ACL}}, pages 8315--8330. Association for Computational Linguistics.

\bibitem[{Sun et~al.(2026{\natexlab{a}})Sun, Yu, Mo, Su, Hui, Liu, and Nie}]{SoG-R1}
Jia~Ao Sun, Hao Yu, Fengran Mo, Zhan Su, Yuchen Hui, Bang Liu, and Jian{-}Yun Nie. 2026{\natexlab{a}}.
\newblock \href {https://doi.org/10.48550/ARXIV.2607.18481} {Search-on-graph-r1: Training large language models to search knowledge graphs with reinforcement learning}.
\newblock \emph{CoRR}, abs/2607.18481.

\bibitem[{Sun et~al.(2024)Sun, Xu, Tang, Wang, Lin, Gong, Ni, Shum, and Guo}]{ToG}
Jiashuo Sun, Chengjin Xu, Lumingyuan Tang, Saizhuo Wang, Chen Lin, Yeyun Gong, Lionel~M. Ni, Heung{-}Yeung Shum, and Jian Guo. 2024.
\newblock \href {https://openreview.net/forum?id=nnVO1PvbTv} {Think-on-graph: Deep and responsible reasoning of large language model on knowledge graph}.
\newblock In \emph{The Twelfth International Conference on Learning Representations, {ICLR} 2024, Vienna, Austria, May 7-11, 2024}. OpenReview.net.

\bibitem[{Sun and Kazakov(2026)}]{Peak-Then-Collapse}
Tianda Sun and Dimitar Kazakov. 2026.
\newblock \href {https://doi.org/10.48550/ARXIV.2605.26037} {Peak-then-collapse and the four interface channels of knowledge-graph tool use}.
\newblock \emph{CoRR}, abs/2605.26037.

\bibitem[{Sun et~al.(2025)Sun, Chen, Zheng, Liu, Wu, Song, Wang, Wang, and Wang}]{KBQA-R1}
Xin Sun, Zhongqi Chen, Xing Zheng, Qiang Liu, Shu Wu, Bowen Song, Zilei Wang, Weiqiang Wang, and Liang Wang. 2025.
\newblock \href {https://doi.org/10.48550/ARXIV.2512.10999} {{KBQA-R1:} reinforcing large language models for knowledge base question answering}.
\newblock \emph{CoRR}, abs/2512.10999.

\bibitem[{Sun et~al.(2026{\natexlab{b}})Sun, Xie, Chen, Liu, Wu, Song, Wang, Wang, and Wang}]{GAPD}
Xin Sun, Jianan Xie, Zhongqi Chen, Qiang Liu, Shu Wu, Bowen Song, Weiqiang Wang, Zilei Wang, and Liang Wang. 2026{\natexlab{b}}.
\newblock \href {https://doi.org/10.48550/ARXIV.2605.29584} {{GAPD:} gold-action policy distillation for agentic reinforcement learning in knowledge base question answering}.
\newblock \emph{CoRR}, abs/2605.29584.

\bibitem[{Talmor and Berant(2018)}]{cwq}
Alon Talmor and Jonathan Berant. 2018.
\newblock \href {https://doi.org/10.18653/V1/N18-1059} {The web as a knowledge-base for answering complex questions}.
\newblock In \emph{Proceedings of the 2018 Conference of the North American Chapter of the Association for Computational Linguistics: Human Language Technologies, {NAACL-HLT} 2018, New Orleans, Louisiana, USA, June 1-6, 2018, Volume 1 (Long Papers)}, pages 641--651. Association for Computational Linguistics.

\bibitem[{Team(2025{\natexlab{a}})}]{kimi}
Kimi Team. 2025{\natexlab{a}}.
\newblock \href {https://doi.org/10.48550/ARXIV.2507.20534} {Kimi {K2:} open agentic intelligence}.
\newblock \emph{CoRR}, abs/2507.20534.

\bibitem[{Team(2025{\natexlab{b}})}]{qwen}
Qwen Team. 2025{\natexlab{b}}.
\newblock \href {https://doi.org/10.48550/ARXIV.2505.09388} {Qwen3 technical report}.
\newblock \emph{CoRR}, abs/2505.09388.

\bibitem[{Touvron et~al.(2023)Touvron, Lavril, Izacard, Martinet, Lachaux, Lacroix, Rozi{\`{e}}re, Goyal, Hambro, Azhar, Rodriguez, Joulin, Grave, and Lample}]{llama}
Hugo Touvron, Thibaut Lavril, Gautier Izacard, Xavier Martinet, Marie{-}Anne Lachaux, Timoth{\'{e}}e Lacroix, Baptiste Rozi{\`{e}}re, Naman Goyal, Eric Hambro, Faisal Azhar, Aur{\'{e}}lien Rodriguez, Armand Joulin, Edouard Grave, and Guillaume Lample. 2023.
\newblock \href {https://doi.org/10.48550/ARXIV.2302.13971} {Llama: Open and efficient foundation language models}.
\newblock \emph{CoRR}, abs/2302.13971.

\bibitem[{Usbeck et~al.(2024)Usbeck, Yan, Perevalov, Jiang, Schulz, Kraft, M{\"{o}}ller, Huang, Reineke, Ngomo, Saleem, and Both}]{qald10en}
Ricardo Usbeck, Xi~Yan, Aleksandr Perevalov, Longquan Jiang, Julius Schulz, Angelie Kraft, Cedric M{\"{o}}ller, Junbo Huang, Jan Reineke, Axel{-}Cyrille~Ngonga Ngomo, Muhammad Saleem, and Andreas Both. 2024.
\newblock \href {https://doi.org/10.3233/SW-233471} {{QALD-10} - the 10th challenge on question answering over linked data: Shifting from dbpedia to wikidata as a {KG} for {KGQA}}.
\newblock \emph{Semantic Web}, 15(6):2193--2207.

\bibitem[{Wang and Yu(2026{\natexlab{a}})}]{KG-Hopper}
Shuai Wang and Yinan Yu. 2026{\natexlab{a}}.
\newblock \href {https://doi.org/10.48550/ARXIV.2603.21440} {Kg-hopper: Empowering compact open llms with knowledge graph reasoning via reinforcement learning}.
\newblock \emph{CoRR}, abs/2603.21440.

\bibitem[{Wang and Yu(2026{\natexlab{b}})}]{KG-Reasoner}
Shuai Wang and Yinan Yu. 2026{\natexlab{b}}.
\newblock \href {https://doi.org/10.48550/ARXIV.2604.12487} {Kg-reasoner: {A} reinforced model for end-to-end multi-hop knowledge graph reasoning}.
\newblock \emph{CoRR}, abs/2604.12487.

\bibitem[{Wang et~al.(2023)Wang, Wei, Schuurmans, Le, Chi, Narang, Chowdhery, and Zhou}]{Self-Consistency}
Xuezhi Wang, Jason Wei, Dale Schuurmans, Quoc~V. Le, Ed~H. Chi, Sharan Narang, Aakanksha Chowdhery, and Denny Zhou. 2023.
\newblock \href {https://openreview.net/forum?id=1PL1NIMMrw} {Self-consistency improves chain of thought reasoning in language models}.
\newblock In \emph{The Eleventh International Conference on Learning Representations, {ICLR} 2023, Kigali, Rwanda, May 1-5, 2023}. OpenReview.net.

\bibitem[{Wei et~al.(2022)Wei, Wang, Schuurmans, Bosma, Ichter, Xia, Chi, Le, and Zhou}]{CoT}
Jason Wei, Xuezhi Wang, Dale Schuurmans, Maarten Bosma, Brian Ichter, Fei Xia, Ed~H. Chi, Quoc~V. Le, and Denny Zhou. 2022.
\newblock \href {http://papers.nips.cc/paper\_files/paper/2022/hash/9d5609613524ecf4f15af0f7b31abca4-Abstract-Conference.html} {Chain-of-thought prompting elicits reasoning in large language models}.
\newblock In \emph{Advances in Neural Information Processing Systems 35: Annual Conference on Neural Information Processing Systems 2022, NeurIPS 2022, New Orleans, LA, USA, November 28 - December 9, 2022}.

\bibitem[{Xu et~al.(2026)Xu, Xu, Liu, Yuan, Peng, Zhao, and Liu}]{GraphWalker}
Shuwen Xu, Yao Xu, Jiaxiang Liu, Chenhao Yuan, Wenshuo Peng, Jun Zhao, and Kang Liu. 2026.
\newblock \href {https://doi.org/10.48550/ARXIV.2603.28533} {Graphwalker: Agentic knowledge graph question answering via synthetic trajectory curriculum}.
\newblock \emph{CoRR}, abs/2603.28533.

\bibitem[{Xu et~al.(2023)Xu, Liu, Culhane, Pertseva, Wu, Semnani, and Lam}]{wikiwebQuestions}
Silei Xu, Shicheng Liu, Theo Culhane, Elizaveta Pertseva, Meng{-}Hsi Wu, Sina~J. Semnani, and Monica~S. Lam. 2023.
\newblock \href {https://doi.org/10.18653/V1/2023.EMNLP-MAIN.353} {Fine-tuned llms know more, hallucinate less with few-shot sequence-to-sequence semantic parsing over wikidata}.
\newblock In \emph{Proceedings of the 2023 Conference on Empirical Methods in Natural Language Processing, {EMNLP} 2023, Singapore, December 6-10, 2023}, pages 5778--5791. Association for Computational Linguistics.

\bibitem[{Yan et~al.(2026)Yan, Chen, Zhou, Yao, Chen, Zhang, Zhang, Zhang, Huang, Duan, and Zhang}]{EoG}
Shiqi Yan, Yubo Chen, Ruiqi Zhou, Zhengxi Yao, Shuai Chen, Tianyi Zhang, Shijie Zhang, Wei~Qiang Zhang, Yongfeng Huang, Haixin Duan, and Yunqi Zhang. 2026.
\newblock \href {https://proceedings.iclr.cc/paper\_files/paper/2026/hash/5d2e24df9cfaad3189833b819c40b392-Abstract-Conference.html} {Explore-on-graph: Incentivizing autonomous exploration of large language models on knowledge graphs with path-refined reward modeling}.
\newblock In \emph{The Fourteenth International Conference on Learning Representations, {ICLR} 2026, Rio de Janeiro, Brazil, April 23-27, 2026}. proceedings.iclr.cc.

\bibitem[{Yao et~al.(2023{\natexlab{a}})Yao, Yu, Zhao, Shafran, Griffiths, Cao, and Narasimhan}]{ToT}
Shunyu Yao, Dian Yu, Jeffrey Zhao, Izhak Shafran, Tom Griffiths, Yuan Cao, and Karthik Narasimhan. 2023{\natexlab{a}}.
\newblock \href {http://papers.nips.cc/paper\_files/paper/2023/hash/271db9922b8d1f4dd7aaef84ed5ac703-Abstract-Conference.html} {Tree of thoughts: Deliberate problem solving with large language models}.
\newblock In \emph{Advances in Neural Information Processing Systems 36: Annual Conference on Neural Information Processing Systems 2023, NeurIPS 2023, New Orleans, LA, USA, December 10 - 16, 2023}.

\bibitem[{Yao et~al.(2023{\natexlab{b}})Yao, Zhao, Yu, Du, Shafran, Narasimhan, and Cao}]{react}
Shunyu Yao, Jeffrey Zhao, Dian Yu, Nan Du, Izhak Shafran, Karthik~R. Narasimhan, and Yuan Cao. 2023{\natexlab{b}}.
\newblock \href {https://openreview.net/forum?id=WE\_vluYUL-X} {React: Synergizing reasoning and acting in language models}.
\newblock In \emph{The Eleventh International Conference on Learning Representations, {ICLR} 2023, Kigali, Rwanda, May 1-5, 2023}. OpenReview.net.

\bibitem[{Yih et~al.(2016)Yih, Richardson, Meek, Chang, and Suh}]{webqsp}
Wen{-}tau Yih, Matthew Richardson, Christopher Meek, Ming{-}Wei Chang, and Jina Suh. 2016.
\newblock \href {https://doi.org/10.18653/V1/P16-2033} {The value of semantic parse labeling for knowledge base question answering}.
\newblock In \emph{Proceedings of the 54th Annual Meeting of the Association for Computational Linguistics, {ACL} 2016, August 7-12, 2016, Berlin, Germany, Volume 2: Short Papers}. The Association for Computer Linguistics.

\bibitem[{Zhang et~al.(2025)Zhang, Jiang, Chi, Chen, Elkoumy, Wang, Wu, Zhou, Pan, Wang, and Ma}]{KGQAGen}
Liangliang Zhang, Zhuorui Jiang, Hongliang Chi, Haoyang Chen, Mohammed Elkoumy, Fali Wang, Qiong Wu, Zhengyi Zhou, Shirui Pan, Suhang Wang, and Yao Ma. 2025.
\newblock \href {http://papers.nips.cc/paper\_files/paper/2025/hash/61a57ab030670598c4200b31100d254c-Abstract-Datasets\_and\_Benchmarks\_Track.html} {Diagnosing and addressing pitfalls in {KG-RAG} datasets: Toward more reliable benchmarking}.
\newblock In \emph{Advances in Neural Information Processing Systems 38: Annual Conference on Neural Information Processing Systems 2025, NeurIPS 2025, San Diego, CA, USA, December 2-7, 2025 / Mexico City, Mexico, November 30 - December 5, 2025}.

\bibitem[{Zheng et~al.(2024)Zheng, Zhang, Zhang, Ye, and Luo}]{zheng2024llamafactory}
Yaowei Zheng, Richong Zhang, Junhao Zhang, Yanhan Ye, and Zheyan Luo. 2024.
\newblock \href {https://doi.org/10.18653/V1/2024.ACL-DEMOS.38} {Llamafactory: Unified efficient fine-tuning of 100+ language models}.
\newblock In \emph{Proceedings of the 62nd Annual Meeting of the Association for Computational Linguistics (Volume 3: System Demonstrations), {ACL} 2024, Bangkok, Thailand, August 11-16, 2024}, pages 400--410. Association for Computational Linguistics.

\end{thebibliography}
\clearpage

\appendix

\section{Proofs}
\label{app:proofs}
Throughout the proofs, a logical form $\ell$ is a conjunctive query: a set of triple patterns over entities and variables with one answer variable, optionally with filters on bound values. For a triple set $\mathcal{S}$, $\operatorname{SPARQL}(\ell,\mathcal{S})$ is the set of values of the answer variable over all assignments that map every pattern of $\ell$ to a triple in $\mathcal{S}$ and satisfy the filters. Such queries are monotone: $\mathcal{S}\subseteq\mathcal{S}'$ implies $\operatorname{SPARQL}(\ell,\mathcal{S})\subseteq\operatorname{SPARQL}(\ell,\mathcal{S}')$. Following Definition 2, $\mathcal{S}$ \emph{supports} $(\ell,y)$ if $y\in\operatorname{SPARQL}(\ell,\mathcal{S})$, and a trajectory $\tau$ is \emph{evidence-covered} if $\operatorname{Cov}(\tau,\mathcal{S}^{*}_{q})=1$, as defined in Equation~\ref{eq:cov}.

\subsection{Proof of Proposition 1}
\label{app:proof1}
\textbf{Proposition 1.} \textit{Backward expansion preserves the answer and yields triples that support it.}\vspace{-2mm}
\begin{proof}
We show by induction that, for every $n\ge 1$, the triple set $\mathcal{S}^{(n)}\subseteq\mathcal{T}$ supports $(\ell^{(n)},y)$.

\textit{Base case.} A seed is retained only if its supporting triples exist in the KG and instantiate $\ell^{(1)}$ with the answer variable bound to $y$. Hence $\mathcal{S}^{(1)}$ supports $(\ell^{(1)},y)$ under some assignment $\mu$.

\textit{Inductive step.} Assume that $\mathcal{S}^{(n)}$ supports $(\ell^{(n)},y)$ under an assignment $\mu$. An expansion selects an entity $u$ in $\ell^{(n)}$ and a verified sub-question with logical form $\tilde{\ell}$, answer variable $x$, and triples $\tilde{\mathcal{S}}\subseteq\mathcal{T}$ that support $(\tilde{\ell},u)$ under an assignment $\tilde{\mu}$ with $\tilde{\mu}(x)=u$. With variables renamed apart, the expanded logical form is
\begin{equation}
    \ell^{(n+1)} = \ell^{(n)}[u\mapsto x]\;\cup\;\tilde{\ell},
\end{equation}
where $[u\mapsto x]$ replaces every occurrence of $u$ with the variable $x$. Consider the assignment $\mu\cup\tilde{\mu}$. Every pattern of $\tilde{\ell}$ is mapped into $\tilde{\mathcal{S}}$, and every pattern of $\ell^{(n)}[u\mapsto x]$ is mapped to the same triple of $\mathcal{S}^{(n)}$ as under $\mu$, since $\tilde{\mu}(x)=u$. The answer variable is still bound to $y$, which gives
\begin{equation}
    y\in\operatorname{SPARQL}\bigl(\ell^{(n+1)},\,\mathcal{S}^{(n+1)}\bigr),
\end{equation}
where $\mathcal{S}^{(n+1)}=\mathcal{S}^{(n)}\cup\tilde{\mathcal{S}}$, so $\mathcal{S}^{(n+1)}$ supports $(\ell^{(n+1)},y)$. Since $\mathcal{S}^{(n+1)}\subseteq\mathcal{T}$, monotonicity also gives $y\in\operatorname{SPARQL}(\ell^{(n+1)},\mathcal{T})$, so the answer is preserved over the full KG. Each expansion replaces one entity with a variable, so $N$ expansions hide $N$ intermediate entities.

The proof guarantees that the accumulated triples are sufficient for the answer, not that they are the only triples that support it. In the special case where the sub-question has answers other than $u$, the result set of $\ell^{(n+1)}$ may grow, but the executability check bounds it by $K_{\text{ans}}$.

In conclusion, every expansion keeps $y$ in the result set over the full KG, and the accumulated triples alone support the expanded logical form. Therefore, backward expansion preserves the answer and yields triples that support it.
\end{proof}

\subsection{Proof of Proposition 2}
\label{app:proof2}
\textbf{Proposition 2.} \textit{ECPR is maximized only by supported answers and rewards each triple once.}\vspace{-2mm}
\begin{proof}
Let $R=R_{\text{out}}+R_{\text{step}}+R_{\text{ecpr}}$ be the composite reward with $\lambda_{\text{ecpr}}>0$, and let $\mathcal{S}^{*}_{q}$ be the supporting triples of a question $q$, which support $(\ell_q,y_q^{*})$ by Proposition 1.

\textit{Supported maximizers.} Since $\operatorname{F1}\le 1$, $R_{\text{step}}\le\lambda_{\text{step}}$, and $\operatorname{Cov}\le 1$, the composite reward is bounded as
\begin{equation}
    R(\tau)\le\lambda_{\text{f1}}+\lambda_{\text{ret}}+\lambda_{\text{step}}+\lambda_{\text{ecpr}},
\end{equation}
where equality holds if and only if every term attains its maximum, in particular only if $\operatorname{Cov}(\tau,\mathcal{S}^{*}_{q})=1$. Assume that an evidence-covered trajectory with a correct answer and only valid actions is feasible within the turn limit. The bound is then attained, so every trajectory $\tau$ that maximizes $R$ satisfies $\mathcal{S}^{*}_{q}\subseteq\mathcal{T}(\tau)$, and monotonicity gives
\begin{equation}
    y_q^{*}\in\operatorname{SPARQL}\bigl(\ell_q,\,\mathcal{T}(\tau)\bigr),
\end{equation}
which means that the answer of $\tau$ is supported.

For outcome rewards alone, $R_{\text{out}}$ in Equation~\ref{eq:rout} depends on $\tau$ only through the prediction $\hat{y}$ and the returned entities $\mathcal{E}(\tau)$. A shortcut trajectory $\tau_s$ that predicts and reaches $y_q^{*}$ with $\operatorname{Cov}(\tau_s,\mathcal{S}^{*}_{q})<1$ therefore attains the maximum $\lambda_{\text{f1}}+\lambda_{\text{ret}}$ of $R_{\text{out}}$. Under the composite reward, it loses at least $\lambda_{\text{ecpr}}\bigl(1-\operatorname{Cov}(\tau_s,\mathcal{S}^{*}_{q})\bigr)>0$ against the bound and is no longer optimal.

\textit{One reward per triple.} Let $U_t=\bigl(\bigcup_{i\le t}\mathcal{T}_i\bigr)\cap\mathcal{S}^{*}_{q}$ be the gold triples retrieved up to step $t$, with $U_0=\emptyset$. The step reward of ECPR is then
\begin{equation}
    r^{\text{ecpr}}_{t}=\lambda_{\text{ecpr}}\,\frac{|U_t|-|U_{t-1}|}{|\mathcal{S}^{*}_{q}|},
\end{equation}
where $U_{t-1}\subseteq U_t$, so $r^{\text{ecpr}}_{t}\ge 0$ and the sum over steps telescopes to $\lambda_{\text{ecpr}}|U_T|/|\mathcal{S}^{*}_{q}|=R_{\text{ecpr}}(\tau)$. If step $t$ returns only triples that were already retrieved, then $U_t=U_{t-1}$ and $r^{\text{ecpr}}_{t}=0$. In contrast, every-match attribution pays $\lambda_{\text{ecpr}}|\mathcal{T}_t\cap\mathcal{S}^{*}_{q}|/|\mathcal{S}^{*}_{q}|$ at each step. Repeating a query that returns $m\ge 1$ gold triples for $T$ steps thus earns $\lambda_{\text{ecpr}}\,Tm/|\mathcal{S}^{*}_{q}|$, which grows with $T$ while the coverage stays at $m/|\mathcal{S}^{*}_{q}|$.

The statement concerns the maximizers of the trajectory-level reward and does not guarantee that policy optimization reaches them. Covering $\mathcal{S}^{*}_{q}$ is sufficient but not necessary for a supported answer, since a question may admit other supporting sets. In the special case where no shortcut trajectory exists for the question, the maximizers of $R_{\text{out}}$ are already evidence-covered, so the two rewards share the same maximizers.

In conclusion, the composite reward attains its maximum only when the retrieved triples support the answer, and first-match attribution never pays a triple twice. Therefore, ECPR is maximized only by supported answers and rewards each triple once.
\end{proof}

\section{FoG Algorithm Detail}
\label{app:redcheck}
\textbf{Redundancy Check.}
To provide a clear overview of the non-redundancy check in Section~\ref{sec:synthesis}, we present its procedure in the form of pseudocode. As shown in Algorithm~\ref{alg:redcheck}, we first build a support graph from the supporting triples and extract the shortest path from each topic entity to the answer node. We then enumerate proper subsets of topic entities in increasing size and execute one SPARQL query per subset. If any subset yields an answer set no larger than that of the full query, the question admits a shortcut and is discarded.

\begin{algorithm}[t]
\small
\caption{Redundancy Check}
\label{alg:redcheck}
\KwIn{\(\mathcal{S}^{(n)}\): supporting triples; \(E_q\): topic entities; \(\mathcal{Y}^{(n)}\): full answer set.}
\KwOut{Redundancy flag \(\textsc{IsRedundant}\in\{\text{True},\text{False}\}\).}

Build an undirected graph \(G\) from \(\mathcal{S}^{(n)}\); label topic-entity nodes with their Wikidata IDs, the answer node with \texttt{?answer}, and intermediates with fresh variables.\par

\For{each \(e \in E_q\)}{
    Compute shortest path \(\mathcal{P}(e)\) in \(G\) from \(e\) to the answer node.\par
    Derive SPARQL triple pattern \(\phi(e)\) from triples along \(\mathcal{P}(e)\).\par
}

\For{\(k = 1\) \textbf{to} \(|E_q|-1\)}{
    \For{each \(Z\subseteq E_q\) with \(|Z|=k\)}{
        \(Q_Z \leftarrow \texttt{SELECT\;?answer\;WHERE\;}\{\bigcup_{e\in Z}\phi(e)\}\)\par
        Execute \(Q_Z\) on the KG \(\rightarrow\) answer set \(\mathcal{Y}^{(n)}_Z\).\par
        \If{\(|\mathcal{Y}^{(n)}_Z|\le|\mathcal{Y}^{(n)}|\)}{
            \Return{\(\textsc{True}\)}
        }
    }
}

\Return{\(\textsc{False}\)}
\end{algorithm}

\textit{Complexity Analysis.}
Given a question with $|E_q|$ topic entities and $|\mathcal{S}^{(n)}|$ supporting triples, extracting the shortest paths by BFS takes $\mathcal{O}(|E_q|\cdot|\mathcal{S}^{(n)}|)$ time. The subset enumeration issues at most $2^{|E_q|}-2$ SPARQL queries and stops at the first redundant subset. Using only the shortest path per topic entity is an approximation, which we audit on 200 candidate questions against an exhaustive enumeration of all path combinations under the same hop limit. The shortest-path check misses a shortcut in 1.5\% of the candidates, while issuing 5.1 rather than 48.6 SPARQL queries per question.

In summary, the redundancy check is dominated by SPARQL execution, and the shortest-path approximation reduces the number of queries by 9.5$\times$ while missing few shortcuts.

\section{Synthetic Data Construction}
\label{app:data}

\subsection{Topic Domains}
\label{app:data_stats}
\begin{figure}[t]
    \centering
    \includegraphics[width=1.0\linewidth]{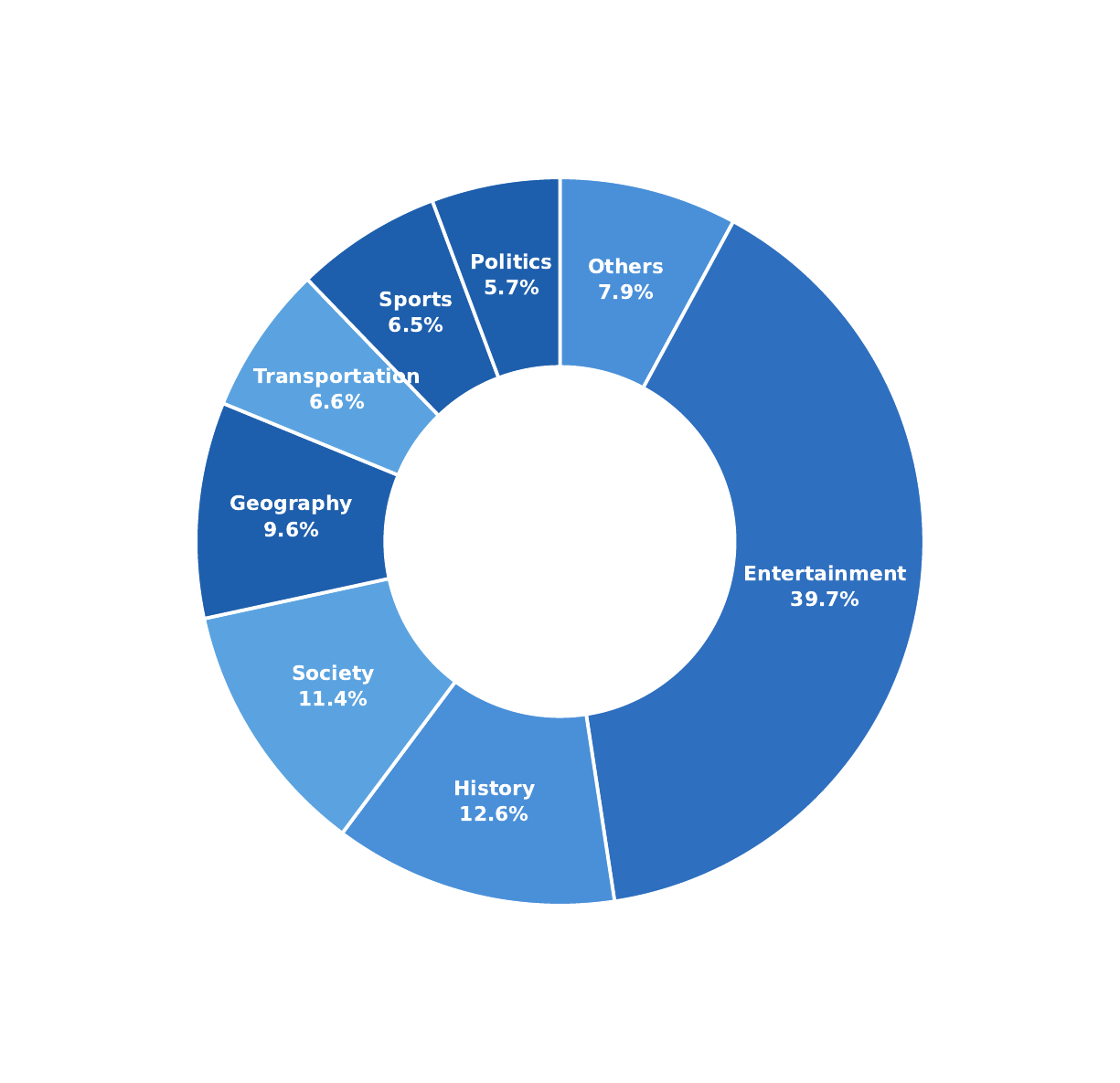}
    \caption{\textbf{Topic-domain distribution of the synthetic questions.} The questions are drawn from major Wikidata domains.}
    \label{fig:data_dist}
\end{figure}
As shown in Figure~\ref{fig:data_dist}, the synthetic questions used for FoG training span seven named Wikidata domains. Entertainment is the largest (39.7\%), reflecting the density of film, television, and music entities in Wikidata, followed by history (12.6\%), society (11.4\%), geography (9.6\%), transportation (6.6\%), sports (6.5\%), and politics (5.7\%); the remaining 7.9\% fall into other domains.

\subsection{Query-Graph Topologies}
\label{app:iso_dist}
\begin{table}[t]
  \centering
  \small
  
  \begin{tabular}{clrr}
  \toprule
  \textbf{Hops} & \textbf{Isomorphism Type} & \textbf{Questions} & \textbf{Proportion} \\
  \midrule
  \multirow{3}{*}{1 hop}
    & \texttt{(1)}           & 500 & 7.59\% \\
    & \texttt{(1)(1)}        & 600 & 9.11\% \\
    & \texttt{(1)(1)(1)}     & 1,211 & 18.38\% \\
  \midrule
  \multirow{4}{*}{2 hop}
    & \texttt{(2)}           & 550 & 8.35\% \\
    & \texttt{(2)(1)}        & 853 & 12.95\% \\
    & \texttt{((1)(1))}      & 700 & 10.63\% \\
    & \texttt{(2)(1)(1)}     & 533 & 8.09\% \\
  \midrule
  \multirow{2}{*}{3 hop}
    & \texttt{(3)}           & 1,086 & 16.48\% \\
    & \texttt{((2)(1))}      & 555 & 8.42\% \\
  \midrule
  \textbf{Total} & - & \textbf{6,588} & \textbf{100.00\%} \\
  \bottomrule
  \end{tabular}
\caption{\textbf{Topology distribution of the synthetic questions.} Query-graph isomorphism types are grouped by hop count, the longest path from a topic entity to the answer.}
\label{tab:distribution}
  \end{table}

To construct questions with diverse reasoning structures, we label each synthetic question with the isomorphism type of its query graph. As shown in Table~\ref{tab:distribution}, the 6,588 questions that pass all checks span nine chain and tree structures from 1 to 3 hops. All verified seeds enter backward expansion, and the 1-hop questions are seeds retained at depth one. Overall, 35.1\%, 40.0\%, and 24.9\% of the questions have 1, 2, and 3 hops. The most frequent types are the three-branch star \texttt{(1)(1)(1)} (18.38\%) and the 3-hop chain \texttt{(3)} (16.48\%), and 67.6\% of the questions have a multi-branch query graph rather than a single chain.

\subsection{Synthesis Cost and Retention}
\label{app:synth_cost}
\begin{table}[t]
\centering
\small
\resizebox{\columnwidth}{!}{
\begin{tabular}{lrrrr}
\toprule
\textbf{Stage} & \textbf{Input} & \textbf{Retained} & \textbf{Rejected} & \textbf{Teacher calls} \\
\midrule
Seed-question generation        & 20,000 & 14,400 & 28.0\% & 20,000 \\
Backward expansion              & 14,400 & 8,784  & 39.0\% & 86,400 \\
Complex-question composition    & 8,784  & 6,588  & 25.0\% & 8,784 \\
\midrule
\textbf{End-to-end}             & 20,000 & 6,588  & 67.1\% & 115,184 \\
\bottomrule
\end{tabular}}
\caption{\textbf{Synthesis cost and retention.}}
\label{tab:synth_cost}
\end{table}

As shown in Table~\ref{tab:synth_cost}, we report the candidates retained, the rejection rate, and the teacher calls at each synthesis stage, all of which use the same teacher model, Kimi-K2~\citep{kimi}. Of 20,000 sampled seed entities, 14,400 yield a verified seed question, 8,784 candidates remain after backward expansion, and 6,588 questions pass all filters, an end-to-end rejection rate of 67.1\%. The pipeline uses 115,184 teacher calls, about 17.5 per retained question, and backward expansion accounts for three quarters of them.

\subsection{Data Leakage Check}
\label{app:leakage}
To ensure that the synthetic questions do not overlap with the test sets, we encode each question with BGE-M3 and compute the cosine similarity between every test question and its most similar synthetic question. The similarity stays below 0.85 for all test questions. On QALD10-en and WikiWebQuestions, no test question exceeds 0.75, and on GTSQA only 6 questions (0.37\%) fall in the 0.80--0.85 range.

\section{KG Interaction Operations}
\label{app:kg_ops}
\begin{table*}[t]
\centering
\small

\resizebox{\textwidth}{!}{         
\begin{tabular}{>{\centering\arraybackslash}m{2.5cm} m{3.8cm} m{5.0cm} m{3.5cm} m{2.2cm}}
\toprule
\textbf{Category} & \textbf{Function} & \textbf{Description} & \textbf{Input Parameters} & \textbf{Output} \\
\midrule
\multirow{2}{2.5cm}{\centering Relation search}
  & \texttt{find\_outgoing\_relations}
  & Given an entity, return all outgoing relations.
  & \texttt{entity\_name}
  & List of outgoing relations \\[4pt]
  & \texttt{find\_incoming\_relations}
  & Given an entity, return all incoming relations.
  & \texttt{entity\_name}
  & List of incoming relations \\
\midrule
\multirow{2}{2.5cm}{\centering Neighborhood expansion}
  & \texttt{get\_tail\_entities}
  & Follow the relation forward from head to tail.
  & \texttt{head\_entity\_name}; \texttt{relation\_name}
  & Set of tail entities \\[4pt]
  & \texttt{get\_head\_entities}
  & Follow the relation backward from tail to head.
  & \texttt{tail\_entity\_name}; \texttt{relation\_name}
  & Set of head entities \\
\midrule
\multicolumn{1}{>{\centering\arraybackslash}m{2.5cm}}{Constraint filtering}
  & \texttt{get\_entity\_by\_constraint}
  & Filter the provided \texttt{entity\_set} and return the entities satisfying the specified constraint.
  & \texttt{entity\_set}; \texttt{relation}; \texttt{operator}; \texttt{value}
  & Entity subset \\
\midrule
\multirow{2}{2.5cm}{\centering Set combination}
  & \texttt{intersect}
  & Return the intersection of the given list of entity sets.
  & \texttt{entity\_set\_list}
  & Intersection set \\[4pt]
  & \texttt{union}
  & Return the union of the given list of entity sets.
  & \texttt{entity\_set\_list}
  & Union set \\
\bottomrule
\end{tabular}}
\caption{\textbf{The seven KG operations of the \texttt{kg-query} action.} Each operation is listed with its description, input parameters, and output.}
\label{tab:kg_operations}

\end{table*}

As shown in Table~\ref{tab:kg_operations}, the \texttt{kg-query} action in Section~\ref{sec:agent} is instantiated with seven operations in four categories: relation search, neighborhood expansion, constraint filtering, and set combination. The table lists the name, description, input parameters, and output of every operation.

\section{Benchmark Datasets}
\label{app:datasets}
\label{app:datasets_detail}
The benchmarks used for FoG evaluation cover two knowledge graphs, with data sourced as follows:

\textbf{WikiWebQuestions:} Ported from WebQuestions and re-annotated with SPARQL queries over Wikidata~\cite{wikiwebQuestions}, consisting of real-world natural language questions.

\textbf{QALD10-en:} Derived from the English subset of the tenth QALD challenge~\cite{qald10en}, the first edition to adopt Wikidata, covering 394 expert-written test questions with multi-hop and qualifier constraints.

\textbf{GTSQA:} Built over Wikidata with 30,144 training and 1,622 test questions~\cite{GTSQA}, each annotated with a gold answer subgraph and a query-graph isomorphism label.

\textbf{WebQSP:} Sourced from Freebase~\cite{webqsp} and used only for the zero-shot cross-KG transfer experiment.

\section{Evaluation Details}
\label{app:evaluation}

\subsection{Evaluation Metrics}
\label{app:metrics}
We evaluate model performance using four complementary metrics that assess two aspects of KGQA: answer accuracy (Hit@1 and F1) and evidence coverage (Cov and RetQ).

\textbf{(i) Hit@1.}
Hit@1 quantifies whether the predicted answer is correct. For each question, we check whether the normalized prediction matches any gold answer, and then average over all evaluated questions:
\begin{equation}
\begin{aligned}
    \mathrm{Hit@1} = \frac{1}{M}\sum_{i=1}^{M}
        \mathbb{I}\bigl[&\exists y \in Y_i^{*}: \\
        &\mathrm{norm}(\hat{y}_i) = \mathrm{norm}(y)\bigr],
\end{aligned}
\end{equation}
where $\hat{y}_i$ and $Y_i^{*}$ denote the prediction and the gold answer set of the $i$-th question, $\mathrm{norm}(\cdot)$ removes superficial differences such as case and punctuation, and $M$ is the total number of evaluated questions in the benchmark.

\textbf{(ii) Token-level F1.}
F1 quantifies the token-level overlap between the predicted and gold answers. For each question, we compute the harmonic mean of token precision and recall, and then average:
\begin{equation}
\small
    \mathrm{F1} = \frac{1}{M}\sum_{i=1}^{M}
        \dfrac{2\cdot\left|\mathrm{tokens}(\hat{y}_i) \cap \mathrm{tokens}(Y_i^{*})\right|}
             {\left|\mathrm{tokens}(\hat{y}_i)\right| + \left|\mathrm{tokens}(Y_i^{*})\right|},
\end{equation}
where $\mathrm{tokens}(\cdot)$ normalizes each answer in a list, splits it on whitespace, and takes the union of the resulting tokens over the list.

\textbf{(iii) Evidence Coverage (Cov).}
Cov quantifies how much of the gold supporting evidence the agent retrieves along its trajectory. For each question, we compute the fraction of gold supporting triples returned by its KG queries, and then average over all questions:
\begin{equation}
    \mathrm{Cov} = \frac{1}{M}\sum_{i=1}^{M}
        \frac{|\mathcal{T}(\tau_i) \cap \mathcal{S}^{*}_{i}|}
             {|\mathcal{S}^{*}_{i}|},
\end{equation}
where $\mathcal{T}(\tau_i)$ denotes the set of triples returned by all \texttt{kg-query} actions in the trajectory $\tau_i$, and $\mathcal{S}^{*}_{i}$ denotes the gold supporting triples of the $i$-th question. This metric reflects whether an answer is backed by the complete evidence path.

\textbf{(iv) Retrieval Grounding (RetQ).}
RetQ quantifies whether the gold answer is retrieved from the KG rather than produced from parametric memory. For each question, we check whether any gold answer appears among the entities returned by its KG queries, and then average over all questions:
\begin{equation}
    \mathrm{RetQ} = \frac{1}{M}\sum_{i=1}^{M}
        \mathbb{I}\!\left[\exists y \in Y_i^{*},\; y \in \mathcal{E}(\tau_i)\right],
\end{equation}
where $\mathcal{E}(\tau_i)$ denotes the set of entities returned by all KG queries in the trajectory $\tau_i$.

\subsection{Gold Supporting Triples}
\label{app:gold_supporting_triples}
The gold supporting triples used for evaluation are distinct from those produced by synthesis, which serve only as ECPR supervision during training. GTSQA provides gold answer subgraphs natively. For WikiWebQuestions, which has no such annotation, we follow EoG~\cite{EoG} and construct them in two stages. We first enumerate candidate paths from each topic entity to each answer entity by BFS under a maximum-hop constraint. An LLM verifier then judges whether each candidate path supports all constraints of the question, and the union of triples on the accepted paths forms the gold supporting triples.

\subsection{Human Validation of Constructed Paths}
\label{app:human_audit}
\begin{table*}[t]
\centering
\small
\begin{tabular}{p{4.2cm} p{5.6cm} p{5.2cm}}
\toprule
\textbf{Question} & \textbf{Candidate path} & \textbf{Reason for rejection} \\
\midrule
Who was the prime minister of Pakistan in 1999?
& Links candidates only to the office ``Prime Minister of Pakistan''.
& Does not establish that the candidate held the office in 1999. \\
\midrule
What town did Justin Bieber grow up in?
& Justin Bieber $\rightarrow$ Canada; Waterloo $\rightarrow$ Canada.
& Sharing the same country does not establish childhood residence. \\
\bottomrule
\end{tabular}
\caption{\textbf{Representative rejected paths.} The verifier checks whether a path covers the semantic constraints of the question, not only whether it reaches a plausible answer entity.}
\label{tab:rejected_paths}
\end{table*}

To validate the constructed paths independently of the LLM verifier, we sample 200 candidate paths from WikiWebQuestions, 100 accepted and 100 rejected by the verifier. Three graduate annotators independently judge whether each path supports all constraints of its question, without seeing the verifier decision. The annotators reach Fleiss' $\kappa=0.91$, and the human--verifier agreement is 93.0\%, with a verifier precision of 0.94 and a recall of 0.92. Table~\ref{tab:rejected_paths} shows two representative rejected paths.

\section{Baseline Details}
\label{app:baselines}
We compare FoG against nine representative baselines that cover the LLM reasoning, retrieval-based, and agent-based paradigms:

\textbf{IO Prompt}~\cite{gpt-4,qwen,deepseekv3} is a KG-free baseline where the LLM directly answers the question from its parametric knowledge. This serves as a lower bound for KG-augmented reasoning.

\textbf{CoT}~\cite{CoT} elicits step-by-step reasoning before the final answer, still without any access to the KG.

\textbf{Self-Consistency}~\cite{Self-Consistency} samples multiple CoT chains and aggregates their answers by majority voting.

\textbf{RoG}~\cite{RoG} generates relation paths as plans, retrieves reasoning paths along them, and answers from the retrieved paths.

\textbf{SubGraphRAG}~\cite{SubGraphRAG} retrieves a subgraph with a lightweight MLP that scores triples in parallel, and feeds it to an LLM reader.

\textbf{GCR}~\cite{GCR} encodes KG paths into a Trie index and constrains LLM decoding with the graph structure.

\textbf{ToG}~\cite{ToG} is a training-free agent that performs beam search over entities and relations through multi-turn KG interaction.

\textbf{PoG}~\cite{PoG} decomposes the question into sub-objectives and explores paths with memory updating and self-reflection.

\textbf{KG-R1}~\cite{KG-R1} trains a single KG agent end-to-end with GRPO under outcome and turn-level action-validity rewards.

\textbf{Adaptation to Wikidata.}
RoG, SubGraphRAG, and GCR were proposed on Freebase and rely on dataset-specific supervision, so they cannot be transferred to a new KG without retraining. We therefore retrain each of them on the training split of every Wikidata benchmark. KG-R1 was also proposed on Freebase. We retrain it on Wikidata with the same training data as FoG.

\label{app:inference_setting}
To ensure fairness, all methods, including FoG, are evaluated under identical conditions with a single inference run per question, and no multi-run aggregation is used. The backbone of each method is listed in Table~\ref{tab:main_result}.

\section{Hyperparameter Settings}
\label{app:training}
\textbf{Supervised Fine-Tuning.}
\label{app:sft}
For SFT, we adopt LLaMA-Factory~\cite{zheng2024llamafactory} and apply LoRA to all linear layers. The training trajectories are distilled from the same teacher model used in synthesis, Kimi-K2~\cite{kimi}, with the prompt in Table~\ref{tab:traj_gen_prompt}. Training runs on a single NVIDIA A100 GPU.

\begin{table}[t]
\centering
\small

\resizebox{\columnwidth}{!}{         
\begin{tabular}{lc}
\toprule
\textbf{Parameter} & \textbf{Value} \\
\midrule
Fine-tuning method       & LoRA \\
LoRA rank / alpha / dropout & 16 / 32 / 0.05 \\
Max sequence length      & 3,000 \\
Training epochs          & 3 \\
Per-device batch size    & 16 \\
Gradient accumulation steps & 8  \\
Learning rate            & $5\times10^{-5}$ \\
LR scheduler             & Cosine, warmup ratio 0.1 \\
Weight decay             & 0.01 \\
Gradient clipping        & 1.0 \\
\bottomrule
\end{tabular}}
\caption{\textbf{Hyperparameters of supervised fine-tuning.}}
\label{tab:sft_hyperparams}
\end{table}

\textbf{Reinforcement Learning.}
\label{app:rl}
For RL, we adopt Verl~\cite{sheng2025hybridflow} and initialize the policy from the SFT checkpoint. Training runs on a single node with 4$\times$NVIDIA A100 (40\,GB) GPUs, using FSDP with parameter and optimizer CPU offloading, and rollouts are executed with vLLM. For each question, the agent samples $N{=}4$ trajectories with at most 10 KG interaction turns. KL regularization uses the $k_3$ estimator with coefficient 0.03, anchored to the SFT model. Table~\ref{tab:sft_hyperparams} and Table~\ref{tab:rl_hyperparams} summarize the detailed hyperparameter configurations used throughout our experiments, and Table~\ref{tab:reward_weights} lists the coefficients of the composite reward in Section~\ref{sec:training}.

\textbf{Advantage Computation.}
\label{app:advantage}
The turn-level advantage in Section~\ref{sec:training} is computed as follows.
\textit{(i) Return.} The return $G_{i,t}$ in Equation~\ref{eq:turn_return} adds the immediate process rewards of turn $t$ to the outcome reward of the trajectory. We use neither discounting nor a reward-to-go, so the evidence reward of a turn is credited to that turn only.
\textit{(ii) Normalization.} The mean $\mu_q$ and standard deviation $\sigma_q$ are computed over all (trajectory, turn) pairs of the $N$ rollouts of the same question. Trajectories of different lengths need no padding, since each contributes as many returns as it has turns.
\textit{(iii) Token mask.} The advantage of a turn is broadcast to the tokens that the policy generates in that turn, i.e., its reasoning and its tool call or answer. Tool observations are part of the context but are masked out of the policy loss and the KL term.
\textit{(iv) Objective.} The policy maximizes
\begin{equation}
\small
\begin{aligned}
&\mathcal{J}(\theta)
= \mathbb{E}_{q,\{\tau_i\}}
\Bigg[
\frac{1}{N}\sum_{i=1}^{N}\frac{1}{|\tau_i|}\sum_{j=1}^{|\tau_i|}
\min \Big(
\rho_{i,j}\hat{A}_{i,t(j)}, \\
&\quad\operatorname{clip}(\rho_{i,j}, 1-\epsilon, 1+\epsilon)\hat{A}_{i,t(j)}
\Big)
\Bigg]
- \beta \,\mathbb{D}_{\mathrm{KL}}(\pi_{\theta}\,\|\,\pi_{\text{ref}}),
\end{aligned}
\end{equation}
where $j$ ranges over the $|\tau_i|$ policy-generated tokens, $t(j)$ is the turn of token $j$, $\rho_{i,j}$ is the probability ratio, and $\pi_{\text{ref}}$ is the SFT model.
\textit{(v) Difference from trajectory-level credit.} When the coverage reward is credited only to the final turn, the turns that actually retrieved the evidence receive the same return as the turns that did not, so the update cannot distinguish them. With first-match attribution, two trajectories with the same coverage still produce different per-turn returns, and the advantage concentrates on the queries that first reach gold triples.

\begin{table}[t]
\centering
\small

\resizebox{\columnwidth}{!}{         
\begin{tabular}{lc}
\toprule
\textbf{Parameter} & \textbf{Value} \\
\midrule
Actor learning rate           & $1\times10^{-6}$ \\
KL loss coefficient           & 0.03 ($k_3$ type) \\
GRPO rollout $N$              & 4 \\
Max interaction turns per question & 10 \\
Train batch size              & 32 \\
PPO mini-batch size           & 16 \\
Max prompt length             & 4,096 \\
Max response length per step  & 8,192 \\
Max observation length        & 1,024 \\
Rollout temperature           & 0.7 \\
Total training steps          & 450 \\
Hardware                      & 4$\times$ NVIDIA A100 (40\,GB) \\
\bottomrule
\end{tabular}}
\caption{\textbf{Hyperparameters of reinforcement learning.}}
\label{tab:rl_hyperparams}
\end{table}

\begin{table}[t]
\centering
\small
\begin{tabular}{llc}
\toprule
\textbf{Reward term} & \textbf{Coefficient} & \textbf{Value} \\
\midrule
Answer F1 ($R_{\text{out}}$)          & $\lambda_{\text{f1}}$   & 0.7 \\
Answer retrieval ($R_{\text{out}}$)   & $\lambda_{\text{ret}}$  & 0.3 \\
Step validity ($R_{\text{step}}$)     & $\lambda_{\text{step}}$ & 0.5 \\
Evidence coverage ($R_{\text{ecpr}}$) & $\lambda_{\text{ecpr}}$ & 1.0 \\
\bottomrule
\end{tabular}
\caption{\textbf{Coefficients of the composite reward.} Each coefficient weights one term of the reward in Section~\ref{sec:training}.}
\label{tab:reward_weights}
\end{table}

\section{Case Study}
\label{app:casestudy}
To better understand how much of the supporting evidence different methods retrieve from the KG, we present a case study on the question: \textit{``Which airline flies the Embraer 195 and is based at the airport named after Margaret Kenyatta's father?''} As shown in Tables~\ref{tab:fogcase}, \ref{tab:togcase}, and~\ref{tab:pogcase}, this question requires resolving the father, the airport named after him, and the airline that is based there and flies the aircraft. All three methods predict the correct answer, but they differ in how much of the evidence chain they recover.

\textbf{FoG} recovers the complete gold evidence. It issues nine KG turns and grounds each hop in the triples returned by the previous one, so every link of the chain can be read from the trajectory.

In contrast, \textbf{ToG} performs four rounds of beam search but recovers only one relevant triple, missing the links for airport naming and airline hub, so its answer relies mainly on parametric knowledge. \textbf{PoG} answers after a single retrieval step, leaving an evidence chain that cannot be audited from the trajectory. This case highlights the strength of FoG in reaching the answer through evidence that is fully retrieved from the KG.

\begin{table*}[ht]
\centering
\footnotesize
\setlength{\tabcolsep}{8pt}
\renewcommand{\arraystretch}{1.3}
\caption{\textbf{Case study of FoG.} The trajectory recovers the complete gold evidence before answering.}

\begin{tabular}{>{\raggedright\arraybackslash}p{0.95\textwidth}}
\rowcolor{gray!20}
\textbf{Question:} Which airline flies the \textbf{Embraer 195} and is based at the airport named after \textbf{Margaret Kenyatta}'s father? \\[2pt]

\rowcolor{blue!10}
\texttt{<think>}I need to find an airline that satisfies two conditions: 1) flies the Embraer 195, and 2) is based at the airport named after Margaret Kenyatta's father. Let me start by exploring the relationships around the first given entity to understand what information is stored about Margaret Kenyatta in the knowledge graph.{</think>} \\

\rowcolor{green!10}
\texttt{<kg-query>} find\_outgoing\_relations(entity\_name=Margaret Kenyatta)  \texttt{</kg-query>} \\
\rowcolor{yellow!10}
\texttt{<information>} Outgoing relations for entity Margaret Kenyatta: country of citizenship, father, given name, instance of, mother, occupation, sex or gender......... \texttt{</information>} \\

\rowcolor{blue!10}
\texttt{<think>} The outgoing triple shows that Margaret Kenyatta has a "father" in the knowledge graph. Let me find out who her father is by querying the triple that directly connects Margaret Kenyatta to her father. {</think>} \\

\rowcolor{green!10}
\texttt{<kg-query>} get\_tail\_entities(head\_entity\_name=Margaret Kenyatta, relation\_name=father)  \texttt{</kg-query>} \\
\rowcolor{yellow!10}
\texttt{<information>} Entities outgoing via relation father from entity Margaret Kenyatta: Jomo Kenyatta.......... \texttt{</information>} \\

\rowcolor{blue!10}
\texttt{<think>} Now I know Jomo Kenyatta is Margaret Kenyatta's father. Next, I need to find the airport named after Jomo Kenyatta. Let me check what relationships exist for Jomo Kenyatta in the knowledge graph to see if there is an airport named after him. {</think>} \\

\rowcolor{green!10}
\texttt{<kg-query>} find\_incoming\_relations(entity\_name=Jomo Kenyatta) \texttt{</kg-query>} \\
\rowcolor{yellow!10}
\texttt{<information>} Incoming relations for entity Jomo Kenyatta: father,named after,spouse.......... \texttt{</information>} \\

\rowcolor{blue!10}
\texttt{<think>} I see that some entities have "named after" pointing to Jomo Kenyatta. Let me find out what entities are named after Jomo Kenyatta in the knowledge graph. {</think>} \\

\rowcolor{green!10}
\texttt{<kg-query>} get\_head\_entities(tail\_entity\_name=Jomo Kenyatta, relation\_name=named after)  \texttt{</kg-query>} \\
\rowcolor{yellow!10}
\texttt{<information>} Entities incoming via relation named after from entity Jomo Kenyatta: Jomo Kenyatta International Airport, Kenyatta International Convention Centre \texttt{</information>} \\

\rowcolor{blue!10}
\texttt{<think>}  Great! I found that Jomo Kenyatta International Airport is named after Jomo Kenyatta. Now I need to find which airline is based at Jomo Kenyatta International Airport. Let me check what outgoing relationships exist for Jomo Kenyatta International Airport to see if there is an airline based there. {</think>} \\

\rowcolor{green!10}
\texttt{<kg-query>} find\_outgoing\_relations(entity\_name=Jomo Kenyatta International Airport)  \texttt{</kg-query>} \\
\rowcolor{yellow!10}
\texttt{<information>} Outgoing relations for entity Jomo Kenyatta International Airport: country, instance of, located in the administrative territorial entity, named after......... \texttt{</information>} \\

\rowcolor{blue!10}
\texttt{<think>} I don't see a direct "airline based at" relationship listed for the airport. Let me check the incoming relationships to see if any airlines have "headquarters location" or similar relationships pointing to this airport. {</think>} \\

\rowcolor{green!10}
\texttt{<kg-query>} find\_incoming\_relations(entity\_name=Jomo Kenyatta International Airport).  \texttt{</kg-query>} \\
\rowcolor{yellow!10}
\texttt{<information>} Incoming relations for entity Jomo Kenyatta International Airport: airline hub \texttt{</information>} \\

\rowcolor{blue!10}
\texttt{<think>} Perfect! I found that some airlines have "airline hub" pointing to Jomo Kenyatta International Airport. Let me find out which airlines have Jomo Kenyatta International Airport as their airline hub. {</think>} \\

\rowcolor{green!10}
\texttt{<kg-query>} get\_head\_entities(tail\_entity\_name=Jomo Kenyatta International Airport, relation\_name=airline hub)  \texttt{</kg-query>} \\
\rowcolor{yellow!10}
\texttt{<information>} Entities incoming via relation airline hub from entity Jomo Kenyatta International Airport: Kenya Airways \texttt{</information>} \\

\rowcolor{blue!10}
\texttt{<think>} Now I have Kenya Airways as the airline based at Jomo Kenyatta International Airport. Finally, I need to verify that Kenya Airways also flies the Embraer 195. Let me check what aircraft types Kenya Airways operates by looking at its outgoing relationships. {</think>} \\

\rowcolor{green!10}
\texttt{<kg-query>} find\_outgoing\_relations(entity\_name=Kenya Airways)  \texttt{</kg-query>} \\
\rowcolor{yellow!10}
\texttt{<information>} Outgoing relations for entity Kenya Airways: airline alliance, airline hub, country, instance of, item operated .........\texttt{</information>} \\

\rowcolor{blue!10}
\texttt{<think>} I see that Kenya Airways has "item operated" as an outgoing relationship. Let me find out what items (likely aircraft) Kenya Airways operates. {</think>} \\

\rowcolor{green!10}
\texttt{<kg-query>} get\_tail\_entities(head\_entity\_name=Kenya Airways, relation\_name=item operated)  \texttt{</kg-query>} \\
\rowcolor{yellow!10}
\texttt{<information>} Entities outgoing via relation item operated from entity Kenya Airways: Boeing 737 Classic,Boeing 737 Next Generation,Boeing 767,Boeing 777,Boeing 787 Dreamliner,Embraer 195 \texttt{</information>} \\

\rowcolor{blue!10}
\texttt{<think>}  I now have enough grounded evidence to give the answer. \texttt{</think>} \\

\rowcolor{green!30}
\texttt{<answer>} ["Kenya Airways"] \texttt{</answer>} \\
\end{tabular}

\label{tab:fogcase}
\end{table*}
\begin{table*}[ht]
\centering
\footnotesize
\setlength{\tabcolsep}{8pt}
\renewcommand{\arraystretch}{1.3}
\caption{\textbf{Case study of ToG.} The trajectory reaches the correct answer while covering only the first part of the gold evidence.}

\begin{tabular}{>{\raggedright\arraybackslash}p{0.95\textwidth}}
\rowcolor{gray!20}
\textbf{Question:} Which airline flies the \textbf{Embraer 195} and is based at the airport named after \textbf{Margaret Kenyatta}'s father? \\[4pt]

\rowcolor{blue!10}
\texttt{Reasoning chains:}

\textbf{Turn 1:}

  `Margaret Kenyatta -> father -> Jomo Kenyatta`;
  `Margaret Kenyatta -> country of citizenship -> Kenya`;
  `Embraer 195 -> subclass of -> E-Jet`;
  `Margaret Kenyatta -> occupation -> diplomat`;
  `Margaret Kenyatta -> occupation -> politician`;
  
\textbf{Turn 2:}

 `Kenya -> named after -> Mount Kenya`;
 `diplomat -> named after -> Diplomatstaden`;
 `E-Jet -> manufacturer -> Embraer`;
 `politician -> category combines topics -> Category:Dutch politicians`;
 `Jomo Kenyatta -> country of citizenship -> Kenya`;
 
\textbf{Turn 3:}

 `Mount Kenya -> country -> Kenya`;
 `Embraer -> headquarters location -> São José dos Campos`;
 `Category:Dutch politicians -> is a list of -> human`;
 `Kenya -> named after -> Mount Kenya`;
 `Diplomatstaden -> located in the administrative territorial entity -> Stockholm Municipality`;
 
\textbf{Turn 4:}

 `Kenya -> named after -> Mount Kenya`;
 `Mount Kenya -> country -> Kenya`;
 `Kenya -> capital -> Nairobi`;
 `São José dos Campos -> country -> Brazil`;
 `São José dos Campos -> located in the administrative territorial entity -> São Paulo`; \\

\rowcolor{red!30}
\texttt{<answer>}First, the airline that flies the Embraer 195 is Kenya Airways. Second, the airport named after Margaret Kenyatta's father is Jomo Kenyatta International Airport. The answer is \{Kenya Airways\}\texttt{</answer>}\\
\end{tabular}

\label{tab:togcase}
\end{table*}
\begin{table*}[ht]
\centering
\footnotesize
\setlength{\tabcolsep}{8pt}
\renewcommand{\arraystretch}{1.3}
\caption{\textbf{Case study of PoG.} The trajectory reaches the correct answer through a shortcut retrieval and misses the bridge evidence.}

\begin{tabular}{>{\raggedright\arraybackslash}p{0.95\textwidth}}
\rowcolor{gray!20}
\textbf{Question:} Which airline flies the \textbf{Embraer 195} and is based at the airport named after \textbf{Margaret Kenyatta}'s father? \\[4pt]

\rowcolor{blue!10}
\texttt{Reasoning chains:}

\textbf{Turn 1:}

  Embraer 195 -> item operated -> Kenya Airways;\\
  
\rowcolor{red!30}
\texttt{<answer>}Kenya Airways operates the Embraer 195 and is based at Jomo Kenyatta International Airport, which is named after Margaret Kenyatta's father. The answer is \{Kenya Airways\}{</answer>}\\
\end{tabular}

\label{tab:pogcase}
\end{table*}

\section{Prompts Used in FoG}
\label{app:prompts}

\subsection{Seed Question Generation Prompt}
As shown in Table~\ref{tab: seedquestion}, this prompt is designed for generating seed questions from a sampled subgraph. It guides the teacher to return a question, its logical form, and its supporting triples, which corresponds to $\pi_{\text{gen}}$ in Equation~\ref{eq:seed}.

\subsection{Backward Expansion Prompt}
As shown in Table~\ref{tab:complex_question_gen_prompt}, this prompt is used by the backward-expansion agent. The agent is instructed to select an anchor entity, retrieve its neighborhood, generate a sub-question whose answer is the anchor, and verify the result, retrying on failure.

\subsection{Trajectory Distillation Prompt}
\label{app:prompts_traj}
As shown in Table~\ref{tab:traj_gen_prompt}, this prompt is used to distill the teacher trajectories for SFT. The teacher is instructed to explore the KG only through the provided tools, explain each call, and answer with entities that appear in the tool responses.

\subsection{Agent Prompt}
\label{app:prompts_train}
As shown in Table~\ref{tab:kgqa_prompt}, this prompt is used by the FoG agent in both training and evaluation. It guides the agent to analyze the question, call the KG operations, and answer from the returned triples.

\begin{table*}[t!]
\centering
\caption{\textbf{Prompt for seed question generation.}}
\label{tab: seedquestion}
 
\begin{minipage}{\textwidth}
\begin{mdframed}[style=outerframe]
\begin{center}
  {\small\textbf{Seed Question Generation Prompt}}
\end{center}
\noindent\rule{\linewidth}{0.4pt}

\noindent{\small\textbf{\textsc{System Prompt}}}

\vspace{4pt}
\begin{mdframed}[style=subframe]

\noindent{\small
You are a question synthesis expert. You are deeply familiar with the structure of the Wikidata knowledge graph and specialize in constructing questions based on given triple information.
}

\vspace{4pt}
\noindent{\small\textbf{Core Concept Definitions}}

\vspace{2pt}
\noindent{\small\textit{Three Entity Roles in this task:}}
\begin{itemize}[leftmargin=*, labelindent=0pt, labelsep=0.6em, itemsep=1pt, topsep=2pt]
  \small
  \item \textbf{Constraint Entity (Con@).} The starting point of the question; an entity the user already knows that must be explicitly mentioned. It anchors the context of the question.
  \item \textbf{Intermediate Entity (Mid@).} The intermediate node in the reasoning chain, obtained through the first reasoning step from the constraint entity. It connects two reasoning steps and must be inferred by the user, never stated explicitly.
  \item \textbf{Answer Entity (Ans@).} The endpoint of the reasoning chain and the final answer requested by the user.
\end{itemize}

\vspace{4pt}
\noindent{\small\textbf{Triple Selection Principles}}
\begin{itemize}[leftmargin=*, labelindent=0pt, labelsep=0.6em, itemsep=1pt, topsep=2pt]
  \small
  \item \textbf{Subset selection.} Select a subset of triples from the given set as the evidence basis. Prefer strong constraints that lead to clear, unambiguous answers.
  \item \textbf{Avoid redundancy.} If a triple already infers B from A, do not select another path that also infers B (no shortcuts allowed).
  \item \textbf{Necessity.} Every selected triple must contribute to the reasoning chain; no irrelevant triples are permitted.
\end{itemize}

\vspace{4pt}
\noindent{\small\textbf{Question Design Principles}}
\begin{itemize}[leftmargin=*, labelindent=0pt, labelsep=0.6em, itemsep=1pt, topsep=2pt]
  \small
  \item \textbf{Information sources.} All information must come exclusively from the selected triples. No external knowledge may be used; the question must be verifiable via a SPARQL query on Wikidata.
  \item \textbf{Reasoning forms.} The question may adopt single-hop reasoning, multi-hop reasoning, intersection queries, union queries, counting, or reverse reasoning.
  \item \textbf{Answer requirements.} The answer must be derived entirely from selected triples and directly verifiable via a SPARQL query.
  \item \textbf{Path guidelines.} A Path is an explanation of the reasoning process, expressed as a reasoning chain describing how the answer is derived.
\end{itemize}

\vspace{4pt}
\noindent{\small\textbf{SPARQL Query Requirements}}
\begin{itemize}[leftmargin=*, labelindent=0pt, labelsep=0.6em, itemsep=1pt, topsep=2pt]
  \small
  \item The generated SPARQL must be syntactically correct and executable on the Wikidata Query Service (WDQS).
  \item Query results must accurately correspond to the answer of the question.
  \item For multi-hop questions, the SPARQL must clearly reflect intermediate variables in the reasoning process.
\end{itemize}

\vspace{4pt}
\noindent{\small\textbf{Entity Usage Rules in Questions}}
\begin{itemize}[leftmargin=*, labelindent=0pt, labelsep=0.6em, itemsep=1pt, topsep=2pt]
  \small
  \item[$\checkmark$] Only constraint entities (question premises) may be explicitly mentioned.
  \item[$\checkmark$] The question must ask for the answer entity (final target).
  \item[$\times$] The specific name of any intermediate entity must \textit{never} appear explicitly in the question.
  \item[$\times$] Intermediate entities must only be implicitly implied through relational wording.
  \item[$\times$] Avoid redundancy: if the answer can be uniquely determined with fewer triples or entities, the question is redundant and must be revised.
\end{itemize}

\end{mdframed}

\vspace{6pt}

\noindent{\small\textbf{\textsc{User Prompt}}}

\vspace{4pt}
\begin{mdframed}[style=subframe]

\noindent{\small\textbf{Relevant Triple Information:}}

\vspace{2pt}
\noindent{\small\texttt{\#\{Relevant triple information\}}}

\end{mdframed}
\end{mdframed}
\end{minipage}
\end{table*}
\begin{table*}[t!]
\centering
\caption{\textbf{Prompt for backward expansion.}}
\label{tab:complex_question_gen_prompt}
 
\begin{minipage}{\textwidth}
\begin{mdframed}[style=outerframe]
 
\begin{center}
  {\small\textbf{Complex Question Generation Prompt}}
\end{center}
\noindent\rule{\linewidth}{0.4pt}
 
\noindent{\small\textsc{System Prompt}}
 
\vspace{4pt}
\begin{mdframed}[style=subframe]
 
\noindent{\small
You are an expert agent specializing in Knowledge Graph Question Answering (KGQA) data synthesis. Select an Anchor Node from a given set of entities and generate a high-quality sub-question that can uniquely identify/locate this anchor entity within the Knowledge Graph.
}
 
\vspace{6pt}
\noindent{\small\textbf{Standard Operating Procedure}:}
\begin{enumerate}[leftmargin=*, labelindent=0pt, labelsep=0.6em, itemsep=1pt, topsep=2pt, label=\textbf{Step \arabic*}]
  \small
  \item  Select an entity as the Anchor Node and call \texttt{retrieve\_entity\_neighborhood} to fetch its neighboring triples.
  \item  Call \texttt{select\_triples\_and\_generate\_subquestion} to choose relevant triples and generate the sub-question.
  \item  Call \texttt{verify\_subquestion} to validate the accuracy and check for redundancy of the generated question.
  \item If validation passes, return the final result in \texttt{<answer>\{{\ldots}\}</answer>} format. If it fails, return to Step~2 and retry.
\end{enumerate}
 
\vspace{6pt}
\noindent{\small\textbf{Strict Constraints:}}
 
\vspace{2pt}
\noindent{\small Prohibited:}
\begin{itemize}[leftmargin=2em, itemsep=1pt, topsep=2pt, label=$\times$]
  \small
  \item \textbf{No shortcuts.} Do not skip any steps or generate the final answer directly.
  \item \textbf{No tool-less answers.} Never return \texttt{<answer>} without having successfully executed the required tool calls.
  \item \textbf{No preconceptions.} Do not assume or hallucinate knowledge before receiving tool outputs.
  \item \textbf{Redundancy rule.} If \texttt{verify\_subquestion} flags the question as ``redundant,'' it means the anchor entity can be uniquely identified using fewer triples or fewer entities in the question. You must retry Step~2.
\end{itemize}
 
\vspace{3pt}
\noindent{\small Required:}
\begin{itemize}[leftmargin=2em, itemsep=1pt, topsep=2pt, label=$\checkmark$]
  \small
  \item \textbf{Sequential execution.} Strictly follow the $1 \to 2 \to 3 \to 4$ workflow.
  \item \textbf{Careful observation.} Analyze the output of each tool before proceeding to the next step.
  \item \textbf{Final approval.} Only output the final result after \texttt{verify\_subquestion} returns a success status.
\end{itemize}
 
\end{mdframed}
 
\vspace{6pt}
 
\noindent{\small\textsc{User Prompt}}
 
\vspace{4pt}
\begin{mdframed}[style=subframe]
 
\noindent{\small\textbf{Output Format:}}
 
\vspace{2pt}
\noindent{\small The final answer must be returned as a JSON object wrapped within \texttt{<answer></answer>} tags:}
 
\vspace{4pt}
\noindent{\small {JSON Format}
}
 
\vspace{2pt}
\noindent{\small If a question cannot be generated, return: \texttt{<answer>do not generate question</answer>}}
 
\vspace{6pt}
\noindent{\small\textbf{Input Entity Information:}}
 
\vspace{2pt}
\noindent{\small
Please select one entity from the list below to serve as the Anchor Node (the answer entity) and generate a sub-question for it:
 
\vspace{2pt}
\texttt{\{entity\_info\}}
}

\end{mdframed}
 
\end{mdframed}
\end{minipage}
 
\end{table*}
\begin{table*}[t!]
\centering
\caption{\textbf{Prompt for trajectory distillation.}}
\label{tab:traj_gen_prompt}
 
\begin{minipage}{\textwidth}
\begin{mdframed}[style=outerframe]
 
\begin{center}
  {\small\textbf{Trajectory distillation Prompt}}
\end{center}
\noindent\rule{\linewidth}{0.4pt}
\noindent{\small\textsc{System Prompt}}
\begin{mdframed}[style=subframe]
 
\noindent{\small
You are a \textbf{Knowledge Graph QA Assistant}. Your task is to explore the knowledge graph \textbf{only by calling the provided function tools}, and answer questions based on the retrieved triplet information.
}
 
\vspace{6pt}
\noindent{\small\textbf{Core Principles} (\textit{Strictly Mandatory}):}
\begin{enumerate}[leftmargin=2em, itemsep=1pt, topsep=2pt, label=\arabic*.]
  \small
  \item \textbf{Tool-first.} You must call at least one tool before giving any answer. Do not answer from memory or guess. If you have not yet received tool results, your next action must be a tool call, not \texttt{<answer>}.
  \item \textbf{No direct answers.} You must not use your own internal knowledge; all information must be obtained from the knowledge graph via function calls.
  \item \textbf{Show reasoning.} Before each function call, briefly explain its purpose; after the call, summarize the information obtained.
  \item \textbf{Evidence-based.} Text inside \texttt{<answer>{\ldots}</answer>} must be only entity name(s) that appeared in the last relevant tool response(s).
  \item \textbf{Step-by-step exploration.} Explore relationships first, then retrieve entities, avoiding blind calls.
\end{enumerate}
 
\vspace{6pt}
\noindent{\small\textbf{Available Functions:} \texttt{\{tool\_call\}}}
 
\vspace{6pt}
\noindent{\small\textbf{Workflow:}}
 
\vspace{2pt}
\noindent{\small\textit{Step 1: Understand the Question.}}
{\small Analyze question type (one-hop, multi-hop, intersection, constraint, hybrid). Identify entities from Entity info. Plan initial exploration.}
 
\vspace{4pt}
\noindent{\small\textit{Step 2: Choose Function Strategy.}}
\begin{itemize}[leftmargin=*, labelindent=0pt, labelsep=0.6em, itemsep=1pt, topsep=2pt]
  \small
  \item Unknown relations: start with \texttt{find\_outgoing\_relations} or \texttt{find\_incoming\_relations}.
  \item Known relation: use \texttt{get\_tail\_entities} or \texttt{get\_head\_entities}.
  \item ``Both A and B'' / ``satisfies both'': get set A, get set B, then \texttt{intersect([set\_A, set\_B])}.
  \item ``Among X, which has attribute = value'': get set X, then \texttt{get\_entity\_by\_constraint}.
\end{itemize}
 
\vspace{4pt}
\noindent{\small\textit{Step 3: Execute Exploration.}}
\begin{enumerate}[leftmargin=2em, itemsep=1pt, topsep=2pt, label=\arabic*.]
  \small
  \item Start with the given entity: call \texttt{find\_outgoing\_relations} or \texttt{find\_incoming\_relations}. Use entity names from Entity info (no QID); use exact relation names from \texttt{find\_*\_relations}.
  \item Based on question, call \texttt{get\_tail\_entities} or \texttt{get\_head\_entities}.
  \item For ``Both A and B'' questions: get set for A, get set for B, then call \texttt{intersect} once.
  \item For constraint questions: (a) obtain candidate set; (b) call \texttt{get\_entity\_by\_constraint(entity\_set, relation, "=", value)} once.
  \item If multi-hop needed, repeat for new entities. Use \texttt{intersect} or \texttt{union} when merging.
\end{enumerate}
 
\vspace{4pt}
\noindent{\small\textit{Step 4: Generate Answer.}}
\begin{itemize}[leftmargin=*, labelindent=0pt, labelsep=0.6em, itemsep=1pt, topsep=2pt]
  \small
  \item As soon as a tool returns a result that uniquely answers the question, output \texttt{<answer>{\ldots}</answer>} immediately; do not call more tools.
  \item Put inside \texttt{<answer>{\ldots}</answer>} \emph{only} entity name(s) that appeared in tool response(s).
  \item If max steps are reached without a clear answer: \texttt{<answer>Cannot find the answer from the knowledge graph.</answer>}
\end{itemize}
 
\end{mdframed}
 
\vspace{6pt}
 
\noindent{\small\textsc{User Prompt}}
 
\vspace{4pt}
\begin{mdframed}[style=subframe]

\vspace{6pt}
\noindent{\small
\textbf{Question:} \texttt{\{question\}}
 
\vspace{3pt}
\textbf{Entity info:} \texttt{\{entity\_info\}}
 
\vspace{6pt}
Please start by \textbf{calling tools}, then use the observations to decide the next step. Answer only after you have tool evidence.
}
 
\end{mdframed}
 
\end{mdframed}
\end{minipage}
 
\end{table*}

\begin{table*}[t!]
\centering
\caption{\textbf{Prompt of the FoG agent in training and evaluation.}}
\label{tab:kgqa_prompt}
 
\begin{minipage}{\textwidth}
\begin{mdframed}[style=outerframe]
 
\begin{center}
  {\small\textbf{FoG Training and Evaluation Prompt}}
\end{center}
\noindent\rule{\linewidth}{0.4pt}
\noindent{\small\textbf{\textsc{System Prompt}}}
 
\vspace{4pt}
\begin{mdframed}[style=subframe]
 
\noindent{\small
You are a knowledge graph question answering assistant. You answer user questions by interacting with a knowledge graph. When a user provides a question and its associated entities, you call tools to query the knowledge graph accordingly. No matter how complex the question is, you will not give up until you believe an answer can be found based on the available information.
}
\vspace{4pt}
\begin{itemize}[leftmargin=*, labelindent=0pt, labelsep=0.6em, itemsep=1pt, topsep=2pt]
  \small
  \item \textbf{Comprehensive analysis.} Combine external triple information obtained through tools to perform analysis and provide a complete answer.
  \item \textbf{Working principles.} Clearly analyze the problem, select appropriate tools, invoke them to obtain information, and ensure the problem is solved efficiently.
\end{itemize}
 
\end{mdframed}
 
\vspace{6pt}
 
\noindent{\small \textbf{\textsc{User Prompt}}}
 
\vspace{4pt}
\begin{mdframed}[style=subframe]
 
\noindent{\small\texttt{<tools>} \small{tools definition} \texttt{</tools>}}
 
\vspace{6pt}
\noindent{\small
The assistant operates in one or more cycles of (thinking $\to$ tool call $\to$ tool response), followed by a final (thinking $\to$ answer) step. All intermediate processes are enclosed within their respective tags. Multiple thinking steps, tool calls, and responses may appear in sequence.
}
 
\vspace{6pt}
\noindent{\small\textbf{Example Response:}}
 
\vspace{3pt}
\begin{lstlisting}[style=inlinecode]
<think> thinking process here </think>
<kg-query>{"name": "tool name", "arguments": {"param": value, ...}}</kg-query>
<information>  tool_response here </information>
 
<think> thinking process here </think>
<kg-query> {"name": "another tool name", "arguments": {...}} </kg-query>
<information> tool_response here </information>
... (more thinking, tool calls, and responses) ...
<think> thinking process here </think>
<answer> answer here </answer>
\end{lstlisting}
 
\vspace{6pt}
\noindent{\small\textbf{User Input:}}
 
\vspace{2pt}
\noindent{\small
\textbf{Question:} \texttt{\{question\}}
 
\vspace{2pt}
\textbf{Entity info}  \texttt{\{entity\_name\}}
}
\end{mdframed}
\end{mdframed}
\end{minipage}
\end{table*}

\end{document}